\documentclass[3p,times]{elsarticle}

\usepackage{ecrc}
\usepackage{algorithm} %format of the algorithm
\usepackage{multirow} %multirow for format of table
\usepackage{amsmath}
\usepackage{xcolor}
\usepackage{amsmath, amssymb, epsfig, graphicx, bm}
\usepackage{cases}
\usepackage{float, subfigure}
\usepackage{color}
\usepackage{epstopdf}
\usepackage{mathrsfs}
\usepackage{amsthm}
\usepackage{amsfonts}
\usepackage{url}
\usepackage[all,cmtip]{xy}
\usepackage{color,hyperref}
\usepackage[noend]{algpseudocode}
\usepackage{algorithmicx,algorithm}
\usepackage{setspace}
\usepackage[misc]{ifsym}

\definecolor{darkblue}{rgb}{0,0,1}
\hypersetup{colorlinks,breaklinks,
	linkcolor=darkblue,urlcolor=darkblue,anchorcolor=darkblue,citecolor=darkblue}

\usepackage[normalem]{ulem} % to use \sout

\newtheorem{theorem}{Theorem}

\newtheorem{lemma}{Lemma}
\newtheorem{remark}{Remark}

\newtheorem{assumption}{Assumption}

\newcommand{\R}{{\mathbb{R}}}

\newcommand{\E}{\mathbb{E}}

\volume{00}

\firstpage{1}

\journalname{ }

\runauth{}

\jid{procs}

\jnltitlelogo{}

\usepackage{amssymb}
\usepackage[figuresright]{rotating}

\begin{document}

	\begin{frontmatter}
		
		
		
		\dochead{}
		
        \title{Byzantine-Robust Decentralized Stochastic Optimization with Stochastic Gradient Noise-Independent Learning Error}


       \author{
       	Jie Peng$^\text{a}$,
       	Weiyu Li$^\text{b}$,
       	Qing Ling$^{\text{a}, }$\footnote{Corresponding author. E-mail address: lingqing556@mail.sysu.edu.cn }
       	% <-this % stops a space
%       	\thanks{Jie Peng and Qing Ling are with School of Computer Science
%       		and Engineering and Guangdong Provincial Key Laboratory of Computational
%       		Science, Sun Yat-Sen University, Guangzhou, Guangdong 510006, China, and
%       		also with Pazhou Lab, Guangzhou, Guangdong 510330, China (Corresponding E-mail: lingqing556@mail.sysu.edu.cn). Weiyu Li is with School of Engineering and Applied Science, Harvard University, Cambridge, MA 02138, USA.}% <-this % stops a space
%       	\thanks{Qing Ling is supported in part by NSF China Grant 61973324, Guangdong Basic and Applied Basic Research Foundation Grant 2021B1515020094, and Guangdong Provincial Key Laboratory of Computational Science Grant 2020- B1212060032. A short preliminary version of this paper has been accepted by IEEE International Conference on Acoustics, Speech and Signal Processing, Singapore, May 22--27, 2022 \cite{peng2022icassp}.}
	}
	 \address{$^\text{a}$ School of Computer Science and Engineering and Guangdong Provincial Key Laboratory of Computationa Science, Sun Yat-Sen University, Guangzhou, Guangdong 510006, China %and also with Pazhou Lab, Guangzhou, Guangdong 510330, China
\\ $^\text{b}$ School of Engineering and Applied Science, Harvard University, Cambridge, MA 02138, USA}

		\begin{abstract}
		This paper studies Byzantine-robust stochastic optimization over a decentralized network, where every agent periodically communicates with its neighbors to exchange local models, and then updates its own local model by stochastic gradient descent (SGD). The performance of such a method is affected by an unknown number of Byzantine agents, which conduct adversarially during the optimization process. To the best of our knowledge, there is no existing work that simultaneously achieves a linear convergence speed and a small learning error. We observe that the learning error is largely dependent on the intrinsic stochastic gradient noise. Motivated by this observation, we introduce two variance reduction methods, stochastic average gradient algorithm (SAGA) and loopless stochastic variance-reduced gradient (LSVRG), to Byzantine-robust decentralized stochastic optimization for eliminating the negative effect of the stochastic gradient noise. The two resulting methods, BRAVO-SAGA and BRAVO-LSVRG, enjoy both linear convergence speeds and stochastic gradient noise-independent learning errors. Such learning errors are optimal for a class of methods based on total variation (TV)-norm regularization and stochastic subgradient update. We conduct extensive numerical experiments to demonstrate their effectiveness under various Byzantine attacks.
		\end{abstract}
		
		
	\end{frontmatter}
	
	% \linenumbers
	
	
\section{Introduction}
%\IEEEPARstart{W}{ith} the rapid increase of data volume and computing power, machine learning technologies have been largely developed in the last decades.  Many problems in machine learning and edge computing \cite{khan2020federated} can be generalized as following optimization problem

\textit{Decentralized stochastic optimization} has attracted immense research interest in recent years, and found various applications in signal processing \cite{srivastava2011constrained,chang2020survey,pu2020survey}, machine learning \cite{lian2017can,assran2019push}, systems control \cite{nedic2009distributed,du2021asynchronous}, communications and networking \cite{giannakis2017decentralized,khan2020federated}, to name a few. Let us consider a decentralized network with $N$ agents. The finite-sum form of the decentralized stochastic optimization problem is given by
\begin{align}\label{problem-decentralized-optimization-1}
	\min_{\tilde{x} \in \R^p} F(\tilde{x}) := \frac{1}{N} \sum_{w=1}^N F_w(\tilde{x}), \quad \text{with} \quad F_w(\tilde{x}) := \frac{1}{J} \sum_{j=1}^J F_{w, j}(\tilde{x}).
\end{align}
%
%with
%%
%\begin{align}\label{problem-decentralized-optimization-2}
%	F_w(\tilde{x}) := \frac{1}{J} \sum_{j=1}^J F_{w, j}(\tilde{x}).
%\end{align}
%%
Here, $\tilde{x} \in \R^p$ is the model to optimize, $F(\tilde{x})$ is the global cost that averages $N$ local costs $F_w(\tilde{x})$, and $F_w(\tilde{x})$ is the local cost of agent $w$ that averages $J$ local sample costs $F_{w, j}(\tilde{x})$. Without loss of generality, we consider that every agent has the same number of samples.  At every iteration, every agent communicates with its neighbors to obtain their local models, aggregates them, and updates its own local model with one or a mini-batch of local samples \cite{tang2018d, mokhtari2016dsa, cui2020momentum}. This is different to \textit{decentralized deterministic optimization} where all the local samples are used in every computation step \cite{shi2015extra,pu2020push, zhang2023accelerated}, and hence fits for scenarios where the number of local samples on every agent is large.

%As one of the important problems in machine learning and cloud computing, decentralized optimization has attracted many interests in recent years. In decentralized optimization, each agent $w$ update its local model $x_w$, which is the local copy of optimization model $\tilde{x}$, by using its local data samples and communicate with the neighbors in each iteration.  After sufficient information exchange, the agents need to finally reach agreement. For the problem formulated by \eqref{problem-decentralized-optimization-1} and \eqref{problem-decentralized-optimization-2}, there exists two types of decentralized methods to solve it. One is for decentralized \textit{deterministic} optimization methods \cite{shi2015extra,pu2020push}, where each agent chooses all local data and computes the full batch gradient in training process. The other is for decentralized \textit{stochastic} optimization methods \cite{tang2018d, mokhtari2016dsa}, where each agent only choose one randomly batch of data and computes stochastic gradient in training process.

However, during the operation of the decentralized system, a number of the agents may become malfunctioning or even behave adversarially due to various uncertainties, such as data corruptions, network failures and external malicious attacks. They can send destructive messages to their neighbors in the communication steps, and hence bias the optimization process. One popular way to describe such uncertainties is the Byzantine attacks model \cite{lamport1982byzantine,yang2020adversary}. The malfunctioning or adversarial agents are termed as Byzantine agents, and the sent destructive messages are termed as Byzantine attacks. This scenario is challenging since the number and identities of the Byzantine agents are unknown, while the Byzantine attacks can be arbitrarily malicious. In this paper, our goal is to develop decentralized stochastic optimization methods that are robust to Byzantine attacks.

\vspace{-0.5em}

\subsection{Related works}

Existing \textit{decentralized deterministic optimization} methods, such as the classical decentralized gradient descent (DGD) \cite{nedic2009distributed}, generally use weighted mean to aggregate neighboring models and are hence vulnerable to Byzantine attacks. The remedy is to replace weighted mean aggregation by robust aggregation. The works of \cite{su2016fault, sundaram2018distributed, su2020byzantine} study Byzantine-robust decentralized deterministic optimization with scalar models, using trimmed mean to replace weighted mean. When the models are vectors, ByRDiE \cite{yang2019byrdie} performs coordinate gradient descent, and applies trimmed mean \cite{su2016fault} to filter out potential malicious elements on every dimension. A centerpoint-based method is proposed in \cite{li2020resilient} to guarantee that the aggregated model always lies in the convex hull of the regular neighbors' models. However, for high-dimensional problems, these two methods are not suitable. Coordinate gradient descent used in ByRDiE is slow, while the centerpoint-based method requires the number of neighbors for every agent to be larger than the problem dimension. BRIDGE \cite{yang2019bridge} extends the work of ByRDiE \cite{yang2019byrdie}, allowing all coordinates to be updated at one time with the help of gradient descent. The work of \cite{kuwaranancharoen2023geometric} provides a general algorithmic framework that includes BRIDGE as a special case, and establishes the linear convergence speed and approximate consensus. In \cite{ben2015robust,xu2018robust}, every agent aggregates the signs of the differences between its own model and the neighbors' such that the influence of Byzantine attacks is limited. The work of \cite{gupta2021byzantine} considers both gradient aggregation and model aggregation. It combines two techniques, comparative gradient elimination to select relatively reliable neighboring gradients, and coordinate-wise trimmed mean to aggregate neighboring models. But \cite{gupta2021byzantine} requires the network to be fully connected.

%Both \cite{yang2019bridge, gupta2021byzantine}  implicitly or explicitly need the $2B$-redundancy property \cite{gupta2020fault, liu2021survey} (Here $B$ is the number of Byzantine agents) which implies that the data across agents are i.i.d., hence not fitting for many real cases since data are often non-i.i.d. in real scenarios. In addition, \cite{ben2015robust} and \cite{xu2018robust} proposes a TV-norm penalization method for Byzantine-robust\textit{ decentralized deterministic}, \textit{decentralized dynamic} optimization problems respectively. However, theses works do not explicitly consider the Byzantine-robust \textit{decentralized stochastic} optimization problem which is of great difficulty and is the major goal of this paper.

%	 [Fault-tolerance...the case of redundancy] propose a important property that one necessary condition of a Byzantine-robust method with $B$-resilience is 2$B$-redundancy which means there exists $B$ agents whose data is redundant. BRIDGE and ByRDiE all needs i.i.d. assumption that the data across the agents are i.i.d. and satisfy the $2B$-redundancy naturally. But the $2B$-redundancy may suffer from noise and is hard to achieve in practical settings. [Approximate Byzantine], which is devoted in Byzantine-robust distributed optimization, propose ($2B, \epsilon$)-redundancy which aims to find a $\epsilon$-approximate optimal solution and is weaker than $2B$-redundancy. Obviously, $2B$-redundancy and ($2B, \epsilon$)-redundancy all imply that the data across partial agents are i.i.d. and redundant.

Although there are an increasing number of methods, such as decentralized parallel stochastic gradient descent (DPSGD) \cite{lian2017can,nedic2009distributed}, for \textit{decentralized stochastic optimization}, few of them consider Byzantine-robustness. Motivated by \cite{ben2015robust} and \cite{xu2018robust}, \cite{peng2021byzantine} introduces a total variation (TV)-norm penalized formulation to force the local models to be close, but allow possible outliers. Applying the stochastic subgradient method to solve this TV-norm penalized formulation yields a Byzantine-robust algorithm, and its sublinear convergence is established.
%Byzantine-robust multi-agent reinforcement learning is studied in \cite{wu2021byzantine}, but the analysis of decentralized policy gradient is different to that of decentralized stochastic gradient. The work of \cite{guo2021byzantine} lets every regular agent evaluate neighboring models with local samples and discard those incurring large losses before aggregation. However, using a large number of local samples for evaluation leads to heavy computation cost, while using a small number causes algorithmic instability. The work of \cite{gorbunov2021secure} proposes a Byzantine-robust algorithm based on secure multi-party computing, but requires that all local samples follow the same data distribution and that the network is fully connected, which are uncommon in decentralized stochastic optimization.
The work of \cite{he2022Byzantine} proposes a robust decentralized centered-clipping rule to aggregate the received local models. The work of \cite{wu2022byzantine} provides a principled guideline for designing proper robust aggregation rules in Byzantine-robust decentralized stochastic optimization. It also proposes iterative outlier scissor as one of such robust aggregation rules. However, the robust aggregation rules introduced in \cite{he2022Byzantine} and \cite{wu2022byzantine} are computationally costly. In addition, the convergence rates established in \cite{he2022Byzantine} and \cite{wu2022byzantine} are both sublinear.

%\red{When all agents have access to a public dataset, \cite{gorbunov2021secure} proposes to randomly choose agents as validators to detect and discard the possible Byzantine agents. But \cite{gorbunov2021secure} has relatively heavy communication and computation burdens due to  the adopted techniques of broadcast channels and multi-party random number generators, and \cite{gorbunov2021secure} only works for homogeneous setting.}

%\cite{peng2021byzantine} considers the TV-norm penalized formulation and proposes to use the stochastic subgradient method to solve it, which achieves approximate Byzantine-robustness.

On the other hand, Byzantine-robust \textit{distributed stochastic optimization} has been a popular topic. Therein, a central server maintains a single global model and aggregates messages from the distributed agents. Most of the methods modify distributed stochastic gradient descent (SGD) by robustly aggregating the stochastic gradients sent from the agents \cite{chen2017distributed,blanchard2017machine}. Such robust aggregation rules include coordinate-wise median, geometric median, trimmed mean, etc.
%\red{Instead of aggregating the  gradients on a central server, \cite{li2019rsa} and \cite{lin2022stochastic} propose aggregating the models of agents. The methods derived from \cite{li2019rsa} and \cite{lin2022stochastic} are capable of effectively handling Byzantine attacks in non-i.i.d. scenarios.}
%\red{Rather than aggregating the stochastic gradients sent from the agents,  proposes to aggregate the model and introduce the TV-norm penalization to control the negative effect of the Byzantine agents.}
Byzantine-robust stochastic second-order methods have also been developed in \cite{cao2020distributed}. To enhance Byzantine-robustness, \cite{cao2019distributed} assumes that the central server has extra clean data, and uses the clean data to distinguish malicious messages from the true stochastic gradients.
However, it is difficult to generalize the above-mentioned works to the decentralized scenario, since there is no central server to collect and aggregate messages in a decentralized network.

%since there exists no central server to collect the global information and all agents only have local information in decentralized optimization.

In stochastic optimization, the noise induced by calculating stochastic gradients plays a critical role and affects convergence. For reducing the negative effect of stochastic gradient noise, variance reduction methods have been employed in \textit{decentralized stochastic optimization} \cite{yuan2019vr,xin2020variance, xin2020decentralized}. In Byzantine-robust \textit{distributed stochastic optimization}, stochastic gradient noise also significantly influences Byzantine-robustness. With large stochastic gradient noise, the regular stochastic gradients are statistically very different, which is conducive for the Byzantine agents to hide themselves and perform attacks \cite{wu2020federated}. In light of this observation, \cite{wu2020federated} proposes to use the stochastic average gradient algorithm (SAGA) to eliminate the impact of inner variation, namely, the stochastic gradient noise caused by the heterogeneity of every regular agent's local samples.
%In addition to lead to a large learning error, the stochastic gradient noise may make the variance between the regular gradients large, which is more conducive to hiding Byzantine gradients %and attacking the system \cite{wu2020federated}. To address this issue, many works cope with the stochastic gradient noise in Byzantine-robust \textit{centralized stochastic} optimization %with the help of variance reduction methods. \cite{wu2020federated} proposes to use SAGA method to eliminate the impact of inner variation which is resulted by the stochastic %noise.
The works of \cite{el2020distributed, karimireddy2021learning} show that momentum acceleration can effectively reduce the impact of stochastic gradient noise and enhance Byzantine-robustness. Aiming at non-convex problems, \cite{khanduri2019byzantine} combines the stochastic variance-reduced gradient (SVRG) method and robust aggregation, and empirically demonstrates the performance. However, none of the existing works considers variance reduction in Byzantine-robust \textit{decentralized stochastic optimization}. Due to the absence of the central server, there is no common model in the decentralized system, and the regular agents have to calculate their stochastic gradients at different local models. Thus, the statistical difference between the regular agents' local models is enlarged, and variance reduction methods are more difficult to analyze.

%The absence of central server makes the variance between agents larger and the variance reduction method is more difficult to analyze in decentralized learning.

Another relevant line of research in the signal processing society is Byzantine-robust decentralized detection. An adaptive threshold is used in \cite{liu2012adaptive} to screen malicious messages and alleviate the negative effect of Byzantine agents. The alternating direction method of multipliers (ADMM) is applied in \cite{kailkhura2017byzantine}, where the trimmed mean aggregation rule is adopted to defend against Byzantine attacks. The work of \cite{kailkhura2016data} considers identifying Byzantine models under data falsification attacks.
%\cite{li2021distributed} is based on distributed voting sequential probability ratio test and prove that the proposed method forms a Nash equilibrium with flip-attack.
For a comprehensive survey, readers are referred to \cite{yang2020adversary}.

\vspace{-0.5em}

\subsection{Our contributions}

Although the methods proposed in \cite{peng2021byzantine, he2022Byzantine, wu2022byzantine} are proven effective in handling Byzantine attacks for decentralized stochastic optimization, the established convergence speeds are sublinear. In particular, for the decentralized Byzantine-robust stochastic aggregation (DRSA) algorithm proposed in \cite{peng2021byzantine}, with a constant step size, it exhibits a linear convergence speed  but the learning error is suboptimal and can be significant (see Section II). With a diminishing step size, DRSA achieves an optimal learning error in order (see Section IV), but only has a sublinear convergence speed. We reveal that the underlying reason of such an unsatisfactory trade-off is the existence of stochastic gradient noise. In this context, our main contributions are as follows.

\noindent \textbf{(C1)} In light of the unsatisfactory trade-off between convergence speed and learning error of the existing methods, this paper proposes BRAVO, \underline{B}yzantine-\underline{R}obust decentralized stochastic optimization \underline{A}ided with \underline{V}ariance reducti\underline{O}n. We introduce two variance reduction methods, SAGA and loopless SVRG (LSVRG), to Byzantine-robust decentralized stochastic optimization for eliminating the negative effect of the stochastic gradient noise. The two methods, BRAVO-SAGA and BRAVO-LSVRG, enjoy both linear convergence speeds and stochastic gradient noise-independent learning errors. We further show that the learning errors are optimal in order for any subgradient-based method that solves the TV-norm penalized decentralized stochastic optimization problem.

%\red{ We provide a lower bound on the learning error, demonstrating that the learning error attained by BRAVO-SAGA and BRAVO-LSVRG matches the lower bound, establishing their optimality among subgradient-based methods when applied to the TV-norm penalized decentralized stochastic optimization problem.}

\noindent \textbf{(C2)} The analysis of BRAVO-SAGA and BRAVO-LSVRG is quite different to that in \cite{peng2021byzantine} due to the applications of the variance reduction techniques. %\blue{To handle the corrected stochastic gradients evaluated at different local models, we introduce two novel Lyapunov functions to enable the establishment of convergence.}
The analysis is also more challenging than those of the Byzantine-robust distributed variance-reduced stochastic optimization methods \cite{wu2020federated}, due to the absence of the central server and the existence of the non-smooth TV-norm penalty term. In addition, \cite{wu2020federated} aggregates all stochastic gradients corrected by variance reduction, while this paper considers aggregation of neighboring models.
%\blue{These challenges are tackled through introducing the new Lyapunov functions, and taking advantages of the bounded subgradients of the TV-norm.}
To tackle the above challenges, we introduce two novel Lyapunov functions that aim to handle the corrected stochastic gradients evaluated at different local models, and take advantages of the bounded subgradients of the TV-norm. %Besides, compared with \cite{wu2020federated}, our works do not assume the heterogeneity of all regular agents is bounded which is more practical.

\noindent \textbf{(C3)} We conduct extensive numerical experiments to demonstrate the effectiveness of the proposed methods.

%\noindent \textbf{(C3)} We conduct extensive numerical experiments to demonstrate the effectiveness of the proposed methods under various Byzantine attacks.

\section{Problem Formulation}

We consider a decentralized undirected network $\mathcal{G} = (\mathcal{V}, \mathcal{E})$ of $N$ agents, in which $\mathcal{V} := \{1, \cdots, N\}$ represents the set of agents and $\mathcal{E}$ represents the set of edges. If $e := (w, v) \in \mathcal{E}$, then agents $w$ and $v$ are neighbors and can communicate with each other. We consider the Byzantine attacks model where a number of agents might be Byzantine and behave arbitrarily. Note that the number and identities of the Byzantine agents are unknown. Denote $\mathcal{R}$ as the set of regular agents with $|\mathcal{R}| = R$ and $\mathcal{B}$ as the set of Byzantine agents with $|\mathcal{B}| = B$. Then, $\mathcal{V} = \mathcal{R} \cup \mathcal{B}$ and $N=R+B$. For agent $w$, denote $\mathcal{R}_w$ and $\mathcal{B}_w$ as the sets of regular neighbors and Byzantine neighbors, respectively.

Since the Byzantine agents can always arbitrarily manipulate their local samples, it is impossible to solve \eqref{problem-decentralized-optimization-1}. Therefore, a reasonable goal is to find a minimizer to the averaged local costs of the regular agents, given by
\begin{align}\label{problem-original}
	\tilde{x}^* = \arg\min_{\tilde{x} \in \R^p} \frac{1}{R} \sum_{w \in \mathcal{R}} F_w(\tilde{x}).
\end{align}
%
%with
%%
%\begin{align}
	%	F_w(\tilde{x}) := \frac{1}{J} \sum_{j=1}^{J} F_{w, j}(\tilde{x})
	%\end{align}
%%
The main difficulty in solving \eqref{problem-original} is that the Byzantine agents will send arbitrarily malicious messages to their neighbors, such that the optimization process is uncontrollable.

Before going further, we make a common assumption in Byzantine-robust decentralized optimization that the network composed of all regular agents is connected \cite{peng2021byzantine}. Intuitively, if one regular agent is surrounded by Byzantine agents, the best model it can expect is the one trained with its own local samples, and can be far away from the true model $\tilde{x}^*$. Denote $\mathcal{E}_R$ as the set of edges that are not attached to any Byzantine agent. The assumption is as follows.
\begin{assumption}\label{assump:network connectivity}
	(\textbf{Network Connectivity}) The network composed of all regular agents, denoted as $(\mathcal{R}, \mathcal{E}_R)$, is connected.
\end{assumption}
%

%Firstly, considering the case that the Byzantine agents are absent.
Let each regular agent $w \in \mathcal{R}$ have a local model $x_w \in \mathbb{R}^p$ and use $x \in \mathbb{R}^{pR}$ to stack all local models of the regular agents. With the connected regular network $(\mathcal{R}, \mathcal{E}_R)$, we can rewrite \eqref{problem-original} to a consensus-constrained form, given by
%	%
%	\begin{align}\label{problem-consensus-constrained}
	%		&\min_{x := [x_i]} \sum_{i \in \mathcal{R}} \E[F(x_i, \xi_i)]. \\
	%		s.&t. \ x_i = x_j, \ \forall i \in \mathcal{R}, \forall j \in \mathcal{R}_i. \nonumber
	%	\end{align}
%	%
%
\begin{align}\label{problem-consensus-constrained}
	\min_{x \in \mathbb{R}^{pR}} ~ & \frac{1}{R} \sum_{w \in \mathcal{R}} F_w(x_w). \\
	\text{s.t.} ~ & \ x_w = x_v, \quad \forall w \in \mathcal{R}, ~ \forall v \in \mathcal{R}_w. \nonumber
\end{align}
Given an optimal solution $\tilde{x}^*$ to \eqref{problem-original}, a longer vector $x^*$ that stacks $R$ vectors $\tilde{x}^*$ is also an optimal solution to  \eqref{problem-consensus-constrained}.

\subsection{Prior work of DRSA}

Following the line of \cite{ben2015robust,xu2018robust,peng2021byzantine}, we penalize the consensus constraints in \eqref{problem-consensus-constrained} by a TV-norm term, which yields
%	%
%	\begin{align}\label{problem-TV}
	%		x^* = \arg\min_{x := [x_i]} \sum_{i \in \mathcal{R}} \left(\E[F(x_i, \xi_i)] + \frac{\lambda}{2} \sum_{j \in \mathcal{R}_i} \|x_i - x_j\|_1 \right)
	%	\end{align}
%	%
%
\begin{align}\label{problem-TV}
	\hspace{-1.6em} x^*=\arg \min_{x \in \mathbb{R}^{pR}} \frac{1}{R} \sum_{w \in \mathcal{R}} \left(F_w(x_w) + \frac{\lambda}{2} \sum_{v \in \mathcal{R}_w} \|x_w - x_v\|_1 \right),
\end{align}
with $\lambda > 0$ being the penalty parameter. The TV-norm penalty term forces the local models to be close, but allows possible outliers. Note that \cite{ben2015robust} and \cite{xu2018robust} consider \textit{decentralized deterministic optimization}, while \cite{peng2021byzantine} investigates \textit{decentralized  stochastic optimization} in the expectation minimization form, not in the finite-sum minimization form here. Nevertheless, the DRSA algorithm in \cite{peng2021byzantine} can still be applied to solve \eqref{problem-TV}.
%Note that compared to the TV-norm penalized approximation formulation in DECEMBER \cite{peng2021byzantine}, we consider the finite-sum optimization rather than the expectation optimization and the local cost function $F_w(x_w)$ includes the regularization term $f_0(x_w)$.  Therefore, the regularization term $f_0(x_w)$ do not exist here.
When the Byzantine agents are absent, applying the stochastic subgradient method to solve \eqref{problem-TV} yields the update
\begin{align}\label{eq-DECEMBER}
	\hspace{-1.1em}	x_w^{k+1} = x_w^k - \alpha^k \left( F'_{w, i_w^k}(x_w^k) + \lambda\sum_{v \in \mathcal{R}_w} sign(x_w^k - x_v^k) \right),
\end{align}
for any regular agent $w \in \mathcal{R}$ at time $k$. Therein, $\alpha^k > 0$ is the step size, and $i_w^k$ is the local sample index (one sample or a mini-batch of samples) selected by regular agent $w$ at time $k$.
%In DECEMBER, each regular agent $w \in \mathcal{R}$ sends its local model to all neighbors and receives the local models of neighbors. Then, each regular agent updates its local model according to \eqref{eq-DECEMBER}.
To run \eqref{eq-DECEMBER}, every regular agent $w \in \mathcal{R}$ needs to collect its neighboring models. However, when the Byzantine agents appear, \eqref{eq-DECEMBER} is not applicable since every regular agent $w \in \mathcal{R}$ does not know the identities of its Byzantine neighbors, while the Byzantine neighbors may send arbitrarily malicious messages for the sake of disturbing the optimization process. Denote $z_v^k \in \mathbb{R}^p$ as an arbitrary model that Byzantine agent $v \in \mathcal{B}$ sends to all neighbors. In fact, it can send different arbitrary models to different neighbors, but for notational convenience we use the same $z_v^k$, which does not affect algorithm development and theoretical analysis. In this case, for regular agent $w \in \mathcal{R}$, the update rule of DRSA is changed from \eqref{eq-DECEMBER} to
\begin{align}\label{eq-DECEMBER-Byzantine-present}
	\hspace{-1em}	x_w^{k+1} = x_w^k - \alpha^k \Bigg( F'_{w, i_w^k}(x_w^k) & + \lambda \sum_{v \in \mathcal{R}_w} sign(x_w^k - x_v^k) + \lambda \sum_{v \in \mathcal{B}_w}sign(x_w^k - z_v^k)\Bigg).
\end{align}

\subsection{Weakness of DRSA}

DRSA is a stochastic subgradient-based method and uses a decreasing step size by default. According to Theorem 2 in \cite{peng2021byzantine}, the theoretical convergence rate of DRSA is sublinear, which is slow. With a proper constant step size, DRSA has a linear convergence rate, but the corresponding learning error is relative to the variance of stochastic gradients, which could be large when the local samples of each regular agent are statistically very different.

\begin{figure}[hb!]
	\centering
	\includegraphics[scale=0.45]{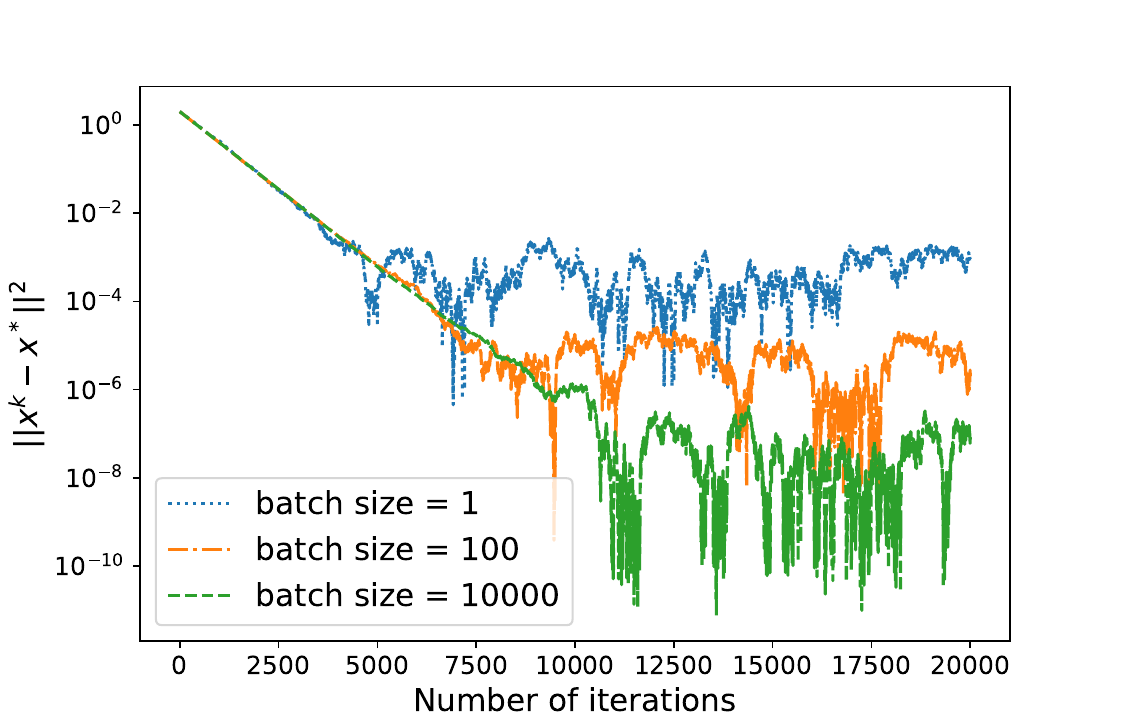}
	\caption{Squared distance between the iterate $x^k$ and the optimal solution $x^*$ under different batch sizes for DRSA. The problem dimension is $p=1$. The network is a complete graph with $N=4$ agents, and every regular agent has $J=10000$ scalar samples $d_{w, j}$ following the standard normal distribution $\mathcal{N}(0, 1)$. The local cost of every regular agent $w \in \mathcal{R}$ is in the least-squares form $F_w(x_w) = \frac{1}{J} \sum_{j=1}^J F_{w, j} (x_w) = \frac{1}{2J} \sum_{j=1}^J (x_w - d_{w, j} )^2$. One of the agents is Byzantine, and performs Gaussian attacks by sending messages following Gaussian distribution $\mathcal{N}(0, 100^2)$. A large batch size corresponds to small stochastic gradient noise. The step size is $\alpha = 0.0008$. The penalty parameter is $\lambda=0.005$.}
	\label{fig:ToyExample}
\end{figure}

Here we illustrate the impact of stochastic gradient noise on DRSA with a simple decentralized stochastic least-squares problem. As shown in Fig. \ref{fig:ToyExample}, when the batch size is small, the stochastic gradient noise is large and the iterate $x^k$ converges to a neighborhood far away from the optimal solution $x^*$. As the batch size increases, the convergence error correspondingly decreases. However, a larger batch size means that more stochastic gradients must be calculated at every time, which is expensive. Inspired by this observation, we propose to utilize computationally cheap variance reduction techniques to reduce the stochastic gradient noise, and in consequence, lower the learning error.

%\begin{example}
	%	Consider a fully connected network with $N = 3$ agents 1, 2 and 3. Let the local cost functions of agent 1, 2 and 3 are $F_1(x_1) = \frac{1}{2} x_1^2 - 50 x_1$, $F_2(x_2) = \frac{1}{2} x_2^2 - 50 x_2$, $F_3(x_3) = \frac{1}{2} x_3^2 - 60 x_3$ respectively. The goal is to solve optimization problem \eqref{problem-consensus-constrained}. First, we consider the case that the stochastic noise exists. In this case, we let the stochastic gradients of agent 1, 2 and 3 are
	%	% $F_1'(x_1) = x_1$, $F_2'(x_2) = x_2$ and $F_3'(x_3) = x_3 - 120$ in iteration $\{2i + 1, i \geq 0\}$, and
	%	%
	%	\begin{align}\label{eq-lsvrg-reference-point}
		% 		F_1'(x_1^k) = \left\{
		%		\begin{aligned}
			%			& = x_1^k, F_2'(x_2^k) = x_2^k, F_3'(x_3^k) = x_3^k - 120  && \ k = 2i (i >= 0), \\
			%			&F_1'(x_1^k) = x_1^k - 100, F_2'(x_2^k) = x_2^k - 100, F_3'(x_3^k) = x_3^k &&\ k = 2i + 1 (i >= 0),
			%		\end{aligned}
		%		\right.
		%	\end{align}
	%	%

	%without incurring much computation overhead.
	
	%\subsection{The case that Byzantine agents exist}
	
	\section{Our Proposed Methods}
	
	In this section, we propose two decentralized stochastic optimization methods equipped with variance reduction techniques to boost Byzantine-robustness, termed as BRAVO-SAGA and BRAVO-LSVRG. Similar to other works on stochastic optimization \cite{mokhtari2016dsa, xin2020variance}, we need to assume that the samplings at the regular agents are independent and uniformly random.
	%\red{Before introduce BRAVO-SAGA and BRAVO-LSVRG, to better describe the sampling procedure in the two methods, we make the following assumption which is commonly used in decentralized variance reduction methods \cite{mokhtari2016dsa, xin2020variance}}
	\begin{assumption}\label{assump:random sampling}
		(\textbf{Independent and Uniformly Random Sampling}) At time $k$, every regular agent $w \in \mathcal{R}$ independently and uniformly randomly select a sample with index $i_w^k \in \{1, \cdots, J\}$.
	\end{assumption}
	
	In BRAVO-SAGA, every regular agent $w \in \mathcal{R}$ maintains a stochastic gradient table in which every item corresponds to a local sample, and updates an item in the table when the corresponding sample is selected to calculate the stochastic gradient. Specifically, let
	\begin{align}\label{eq-saga-gradient-table}
		\phi^{k+1}_{w, j} = \left\{
		\begin{aligned}
			&\phi^k_{w, j}, && j \neq i_w^k, \\
			&x_w^k, && j = i_w^k,
		\end{aligned}
		\right.
	\end{align}
	where
	%\red{$i_w^k$ denotes the index of the independently and uniformly randomly selected sample,}
	$\phi^{k}_{w, j}$ represents the most recent local model used in calculating $F'_{w, j}$ on agent $w$, prior to time $k$. If the selected sample index $i_w^k = j$ at time $k$, then $\phi^{k+1}_{w, j} = x_w^k$; otherwise $\phi^{k+1}_{w, j} = \phi^k_{w, j}$. Therefore, $F'_{w, j}(\phi^k_{w, j})$ refers to the most recent stochastic gradient for sample $j$ on agent $w$, prior to time $k$.
	
	Rather than using a stochastic gradient $F'_{w, i_w^k}(x_w^k)$ to update its local model, here every regular agent $w \in \mathcal{R}$ calculates a corrected stochastic gradient $g_w^k$, which is defined as
	\begin{align}\label{eq-saga}
		g_w^k = F'_{w, i_w^k}(x_w^k) - F'_{w, i_w^k}(\phi_{w, i_w^k}^k) + \frac{1}{J}\sum_{j=1}^{J} F'_{w, j}(\phi_{w, j}^k).
	\end{align}
	We can observe that the corrected stochastic gradient $g_w^k$ modifies the stochastic gradient $F'_{w, i_w^k}(x_w^k)$ by subtracting the stored one $ F'_{w, i_w^k}(\phi_{w, i_w^k}^k)$ that also corresponds to sample index $i_w^k$, and adding the average $\frac{1}{J}\sum_{j=1}^{J} F'_{w, j}(\phi_{w, j}^k)$ of all stored stochastic gradients. Such a variance reduction operation works on the local samples and is irrelevant with the TV-norm penalty. Thus, we replace the original stochastic gradient $F'_{w, i_w^k}(x_w^k)$ in \eqref{eq-DECEMBER-Byzantine-present} by $g_w^k$,  and write the update rule of every regular agent $w \in \mathcal{R}$ in BRAVO-SAGA as
	%%
	%\begin{align}\label{eq-DECEMBER-SAGA}
		%	x_w^{k+1} = x_w^k - \alpha \cdot \left(g_w^k + \lambda \sum_{v \in \mathcal{R}_w} sign(x_w^k - x_v^k) \right)
		%\end{align}
	%%
	%
	\begin{align}\label{eq-DECEMBER-VR}
		x_w^{k+1} = x_w^k - \alpha \Bigg(g_w^k & + \lambda \sum_{v \in \mathcal{R}_w} sign(x_w^k - x_v^k) + \lambda \sum_{v \in \mathcal{B}_w}sign(x_w^k - z_v^k)\Bigg).
	\end{align}
	It is worth noting that we use a constant step size $\alpha > 0$ here, taking advantage of the variance reduction property of SAGA \cite{defazio2014saga}. In Section \ref{sec:Theoretical Analysis}, we will show that BRAVO-SAGA can linearly converge with a constant step size.

	\begin{algorithm}
		\caption{BRAVO-SAGA} {\bf Input:} $x_w^0 \in \R^p$ and $F'_{w, j}(\phi_{w, j}^0) = F'_{w, j}(x_w^0), j = 1, \cdots, J$, $\forall w \in \mathcal{R}$. $\lambda >0$ and $\alpha > 0$.
		\begin{algorithmic}[1]
			\For{$k = 0, 1, \cdots$, every regular agent $w \in \mathcal{R}$}
			\State Broadcast its current model $x_w^k$ to all neighbors.
			\State Receive $x_v^k$ from regular neighbors $v \in \mathcal{R}_w$ and $z^k_v$ from Byzantine neighbors $v \in \mathcal{B}_w$.
			\State Sample $i_w^k$ from $ \{1, \cdots, J\}$.
			\State Store stochastic gradient $F'_{w, i_w^k}(\phi_{w, i_w^k}^k) =  F'_{w, i_w^k}(x_w^k)$.
			\State Update $g_w^k$ according to \eqref{eq-saga}.
			\State Update local iterate $x^{k+1}_w$ according to \eqref{eq-DECEMBER-VR}.
			\EndFor \State \textbf{end for}
		\end{algorithmic}
	\end{algorithm}
	\begin{algorithm}
		\caption{BRAVO-LSVRG} {\bf Input:} $x_w^0 \in \R^p$ and  $y_w^0 = x_w^0$, $\forall w \in \mathcal{R}$. $\lambda >0$ and $\alpha > 0$.
		\begin{algorithmic}[1]
			\For{$k = 0, 1, \cdots$, every regular agent $w \in \mathcal{R}$}
			\State Broadcast its current model $x_w^k$ to all neighbors.
			\State Receive $x_v^k$ from regular neighbors $v \in \mathcal{R}_w$ and $z^k_v$ from Byzantine neighbors $v \in \mathcal{B}_w$.
			\State Sample $i_w^k$ from $ \{1, \cdots, J\}$.
			\State With probability $1 / J$, $y_w^{k+1} = x_w^k$ and calculate local
			\State \quad full gradient; otherwise, $y_w^{k+1} = y_w^k$.
			\State Update $g_w^k$ according to \eqref{eq-lsvrg}.
			\State Update local iterate $x^{k+1}_w$ according to \eqref{eq-DECEMBER-VR}.
			\EndFor \State \textbf{end for}
		\end{algorithmic}
	\end{algorithm}

	Inheriting the property of SAGA, BRAVO-SAGA requires every regular agent to maintain a stochastic gradient table, which brings extra storage cost. For storage-limited applications, we can replace SAGA by loopless SVRG (as known as LSVRG) \cite{kovalev2020don}, which incurs no extra storage cost, although doubles the computation cost per iteration.

	%LSVRG has been proven superior than SVRG that uses two loops \cite{kovalev2020don}.
	
	Define
	\begin{align}\label{eq-lsvrg-reference-point}
		y_w^{k+1} = \left\{
		\begin{aligned}
			&x_w^k, &&\text{with probability} \  \frac{1}{J}, \\
			&y_w^k, &&\text{with probability} \ 1 - \frac{1}{J},
		\end{aligned}
		\right.
	\end{align}
	where $y_w^{k+1}$ is a reference point at which the full gradient $F'_{w}(x_w)$ is calculated. In BRAVO-LSVRG, every regular agent $w \in \mathcal{R}$ also updates with \eqref{eq-DECEMBER-VR}, where the corrected stochastic gradient is defined as
	%%
	%\begin{align}\label{eq-lsvrg}
		%	g_w^k = F'_{w, i_w^k}(x_w^k) - F'_{w, i_w^k}(y_w^k) + \frac{1}{J} \sum_{j=1}^J  F'_{w, j}(y_w^k),
		%\end{align}
	%%
	%
	\begin{align}\label{eq-lsvrg}
		g_w^k = F'_{w, i_w^k}(x_w^k) - F'_{w, i_w^k}(y_w^k) +  F'_{w}(y_w^k).
	\end{align}
	Observe that at every time, with probability $\frac{1}{J}$, $y_w^{k+1}$ is updated to $x_w^k$, meaning that the local full gradient must be calculated. By doubling the number of the stochastic gradient calculations per iteration (in expectation), BRAVO-LSVRG avoids the high storage cost of BRAVO-SAGA.

	%For DECEMBER-SAGA and DECEMBER-LSVRG, the update rule becomes
	%%
	%\begin{align}\label{eq-DECEMBER-VR}
		%	x_w^{k+1} = &x_w^k - \alpha \cdot \Bigg(g_w^k \nonumber\\
		%	&+ \lambda \sum_{v \in \mathcal{R}_w} sign(x_w^k - x_v^k) + \lambda \sum_{v \in \mathcal{B}_w}sign(x_w^k - z_v^k)\Bigg)
		%\end{align}
	%%
	%where $g_w^k$ is defined in \eqref{eq-saga} for DECEMBER-SAGA and defined in \eqref{eq-lsvrg} for DECEMBER-LSVRG.
	
	The two proposed methods are outlined in Algorithm 1 and Algorithm 2, respectively.

	\vspace{-0.2em}

	\section{Theoretical Analysis}
	\label{sec:Theoretical Analysis}
	In this section, we theoretically analyze the performance of the proposed BRAVO-SAGA and BRAVO-LSVRG.
	
	\vspace{-0.5em}
	
	\subsection{Assumptions}
	We make the following assumptions that are common in the analysis of decentralized stochastic optimization \cite{lian2017can, peng2021byzantine} and variance reduction \cite{defazio2014saga, kovalev2020don}.
	\begin{assumption}\label{assump:strong convexity}
		\textbf{(Strong Convexity)} For any model $\tilde{x} \in \R^p$ and every regular agent $w \in \mathcal{R}$, the local cost $F_w(\tilde{x})$ is strongly convex with constant $\mu_w$.
	\end{assumption}
	\begin{assumption}\label{assump:lipschitz continuous gradients}
		\textbf{(Lipschitz Continuous Gradients)} For any model $\tilde{x} \in \R^p$ and every regular agent $w \in \mathcal{R}$, the local sample cost $F_{w, i_w^k}(\tilde{x})$ has Lipschitz continuous gradients with constant $L_w$.
	\end{assumption}
	\begin{assumption}\label{assump:bounded variance}
		\textbf{(Bounded Variance)} For any model $\tilde{x} \in \R^p$, the variance of $F'_{w, i_w^k}(\tilde{x})$ is upper bounded by $\delta_w^2$, namely, $\E_{i_w^k}\|F'_{w, i_w^k}(\tilde{x}) -  F'_{w}(\tilde{x})\|^2 \leq \delta_w^2$.
	\end{assumption}

	Assumption \ref{assump:bounded variance} bounds the variance of the local stochastic gradients. It is worth noting that BRAVO-SAGA and BRAVO-LSVRG only need Assumptions \ref{assump:strong convexity} and \ref{assump:lipschitz continuous gradients}, but do not need Assumption \ref{assump:bounded variance} that is necessary for the theoretical analysis of DRSA \cite{peng2021byzantine}. The reason is that the adopted variance reduction methods are able to eliminate the effect of stochastic gradient noise. We do not assume homogeneity of regular agents; their local costs can be totally different.
	
	%\red{\subsection{Lower bound on Convergence error}
		%
		%Note that the update rule of regular agent $w \in \mathcal{R}$ in  DRSA, BRAVO-SAGA and BRAVO-LSVRG can be formulated with the same expression as follows
		%%
		%%
		%\begin{align}\label{eq-Framework}
			%	x_w^{k+1} = x_w^k - \alpha \Bigg(h_w^k & + \lambda \sum_{v \in \mathcal{R}_w} sign(x_w^k - x_v^k) \\
			%	& + \lambda \sum_{v \in \mathcal{B}_w}sign(x_w^k - z_v^k)\Bigg). \nonumber
			%\end{align}
		%%
		%%
		%where the stochastic gradient $h_w^k$ is an unbiased estimator of the full gradient which means $\E[h_w^k] = F'_w(x_w^k)$. \eqref{eq-Framework} represents the basic algorithm to solve the TV-norm penalized optimization problem \eqref{problem-TV}.
		%}
	
	\vspace{-0.5em}
	
	\subsection{Main theorems}
	As the network topology plays a critical role in the analysis, we define $A \in \R^{R \times |\mathcal{E}_R|}$ as the oriented agent-edge incidence matrix of $(\mathcal{R}, \mathcal{E}_R)$. To be specific, for an edge $e =(w, v) \in \mathcal{E}_R$ with $w < v$, the $(w ,e)$-th entry of $A$ is 1 and the $(v, e)$-th entry of $A$ is $-1$. We review the following lemma.
	\begin{lemma}\label{thm1}
		(\cite{peng2021byzantine}, Theorem 1) Suppose that Assumptions \ref{assump:network connectivity} and \ref{assump:strong convexity} hold true. If $\lambda \geq \lambda_0 := \frac{\sqrt{R}}{\tilde{\sigma}_{\min}(A)} \max_{w \in \mathcal{R}} \| F'_w(\tilde{x}^*)\|_{\infty}$ where $\tilde{\sigma}_{\min}(A)$ is the minimum nonzero singular value of A, then for the optimal solution $x^*$ of \eqref{problem-TV} and the optimal solution $\tilde{x}^*$ of \eqref{problem-original}, we have that $x^*$ stacks $R$ vectors $\tilde{x}^*$ as $[\cdots; \tilde{x}^*; \cdots]$.
	\end{lemma}
	%
	
%	\red{
%		It is important to note that the critical value, denoted as $\lambda_0$, is directly influenced by the heterogeneity across the regular agents. This implies that the penalty parameter $\lambda$ is also related to the heterogeneity across the regular agents. Specifically, when there is significant heterogeneity across the regular agents, both the critical value $\lambda_0$ and the penalty parameter $\lambda$ increase accordingly.}

Since BRAVO-SAGA, BRAVO-LSVRG and DRSA solve the same TV-norm penalized problem \eqref{problem-TV}, they share Lemma 1, which shows that \eqref{problem-TV} is equivalent to the original problem \eqref{problem-original} when the penalty parameter $\lambda$ is sufficiently large. Note that the threshold $\lambda_0$ is determined by the heterogeneity of the regular agents' local costs. Higher heterogeneity leads to larger $\lambda_0$. Below we only consider the regime of large $\lambda$ and focus on the convergence to \eqref{problem-TV}.
	
	The next theorem demonstrates that both BRAVO-SAGA and BRAVO-LSVRG can achieve linear convergence, and the learning errors are bounded.
	
	\begin{theorem}\label{thm2}
		Suppose Assumptions \ref{assump:network connectivity}, \ref{assump:random sampling}, \ref{assump:strong convexity}, \ref{assump:lipschitz continuous gradients} hold true. Denote $\eta = \min_{w \in \mathcal{R}} \{ \frac{\mu_w L_w}{\mu_w + L_w} \} - \frac{\epsilon}{2} > 0$, $\epsilon \in (0, \min_{w \in \mathcal{R}} \{\frac{2 \mu_w L_w}{\mu_w + L_w}\})$, and $L = \max_{w \in \mathcal{R}} L_w$. Set the step size of BRAVO-SAGA and BRAVO-LSVRG as $\alpha \leq \frac{\eta}{12L^2J}.$
%		%
%		\begin{align}\label{th2-1-stepsize}
%			\alpha \leq \frac{\eta}{12L^2J}.
%		\end{align}
%		%
		We have
		\begin{align}\label{th2-2}
			\E[V^k] \leq (1 - \eta\alpha)^k V^0 + \Delta,
		\end{align}
		where
		\begin{align}\label{th2-3}
			V^k := \|x^k - x^*\|^2 + 8J\alpha^2 L^2 S^k,
		\end{align}
		\vspace{-50pt}
		\begin{align}\label{th2-4}
			\Delta :=  \frac{\alpha}{\eta}\sum_{w \in \mathcal{R}} (32\lambda^2|\mathcal{R}_w|^2p + 4\lambda^2|\mathcal{B}_w|^2p)  + \frac{1}{\epsilon\eta}\sum_{w \in \mathcal{R}}\lambda^2|\mathcal{B}_w|^2p.
		\end{align}
		For BRAVO-LSVRG, we define $S^k$ as
		\begin{align}\label{th2-5}
			S^k := \sum_{w \in \mathcal{R}} \|y_w^k - x_w^*\|^2.
		\end{align}
		For BRAVO-SAGA, we define $S^k$ as
		\begin{align}\label{th2-6}
			S^k := \sum_{w \in \mathcal{R}}\frac{1}{J}\sum_{j=1}^J\|\phi_{w, j}^k - x_w^*\|^2.
		\end{align}
		In \eqref{th2-2}, the expectation is taken over all random variables.
	\end{theorem}
	The difference between \eqref{th2-5} and \eqref{th2-6} comes from the ways of SAGA and LSVRG to correct the stochastic gradients. Thus, we choose different Lyapunov functions $V^k$ and derive a unified convergence expression for the two methods. Theorem \ref{thm2} asserts that {BRAVO-SAGA and BRAVO-LSVRG can linearly converge to a neighborhood of the optimal solution $x^*$ of \eqref{problem-TV}; the size of the neighborhood is determined by the penalty parameter $\lambda$, the summation of the squared numbers of Byzantine neighbors $\sum_{w \in \mathcal{R}} |\mathcal{B}_w|^2$, the summation of the squared numbers of regular neighbors $\sum_{w \in \mathcal{R}} |\mathcal{R}_w|^2$, and the problem dimension $p$.

		\begin{remark}
			According to Theorem \ref{thm2}, we know that the step size in BRAVO-SAGA and BRAVO-LSVRG is quite important to their performance. If we choose a small step size, the learning error $\Delta$ is small but the convergence speed is relatively slow. Otherwise, if we choose a large step size, the convergence speed will be fast but the learning error is large. One can balance the two performance metrics through judiciously tuning the step size.
		\end{remark}

			Next, we show that the learning errors established in Theorem \ref{thm2} are optimal in order for any subgradient-based method that solves the TV-norm penalized decentralized stochastic optimization problem in \eqref{problem-TV}. The class of such methods satisfy the following assumption.

			\begin{assumption}\label{assump:span condition}
				\textbf{(Span Condition)} The iterative method solving \eqref{problem-TV} generates the sequence of regular agents' models $\{x_w^k\}_{w \in \mathcal{R}}$ via
				\begin{align}\label{eq-subgradient-based}
					x_w^k \in x_w^0 + \text{span} \{\hat{h}_w^0, \hat{h}_w^1, \cdots, \hat{h}_w^{k-1}\}, \quad \forall w \in \mathcal{R},
				\end{align}
				where $\hat{h}_w^k:= h_w^k + \lambda \sum_{v \in \mathcal{R}_w} sign(x_w^k - x_v^k) + \lambda \sum_{v \in \mathcal{B}_w} sign(x_w^k - z_v^k)$ is a stochastic subgradient, while $h_w^k$  is a  linear combination of $\{F'_{w, 1}(x_w^0), \cdots, F'_{w, J}(x_w^0), F'_{w, 1}(x_w^1), \cdots, F'_{w, J}(x_w^1), \cdots, F'_{w, 1}(x_w^k), \cdots, F'_{w, J}(x_w^k)\}$ and satisfies $\E[h_w^k] = F'_w(x_w^k)$.
			\end{assumption}
			Such a span condition is standard in analyzing the performance bounds of (stochastic) first-order methods \cite{nesterov2003introductory, lan2018optimal}. Methods satisfying Assumption \ref{assump:span condition} include but are not limited to DRSA, BRAVO-SAGA and BRAVO-LSVRG.

			\begin{theorem} \label{thm-bound}
			Given any decentralized stochastic optimization method to solve \eqref{problem-TV} that: (i) satisfies Assumptions \ref{assump:random sampling} and \ref{assump:span condition}; (ii) initializes the regular agents' models at the same point, we can: (i) find $R$ local costs $\{F_1(x),  \cdots, F_R(x)\}$ satisfying Assumption \ref{assump:strong convexity} for the regular agents, among which each local cost $F_{w}(x)$ is composed of the local sample costs $\{F_{w, 1}(x), \cdots,  F_{w, J}(x)\}$ satisfying Assumption \ref{assump:lipschitz continuous gradients}; (ii) find a network topology satisfying Assumption \ref{assump:network connectivity}, within which each regular agent $w$ has $\mathcal{B}_w$ Byzantine neighbors, such that the learning error of this method is at least
			\begin{align}\label{eq-lower-bound}
				\E \|x^k - x^*\|^2 = \Omega( \sum_{w \in \mathcal{R}}\lambda^2|\mathcal{B}_w|^2 p).
			\end{align}
			%
			%Here the lower bound \eqref{eq-lower-bound} omits the  constants $\eta, \epsilon, \alpha$ which remain unchanged across various Byzantine attacks scenarios.
			\end{theorem}

The lower bound on the learning error, as given in \eqref{eq-lower-bound}, is influenced by the heterogeneity across the local costs of regular agents (represented by the penalty parameter $\lambda$) and the impact of Byzantine agents (represented by the term $\sum_{w \in \mathcal{R}} |\mathcal{B}_w|^2$). Note that the lower bound given in \cite{karimireddy2021byzantine} for Byzantine-robust distributed stochastic optimization is also determined by the heterogeneity across the local costs of regular agents and the impact of Byzantine agents.

		\begin{remark}
Although the Lyapunov functions in Theorem \ref{thm2} and Theorem \ref{thm-bound} are different, the learning errors are still comparable. Observe that the learning error in Theorem \ref{thm2} is optimal in order when the summation of the squared numbers of Byzantine neighbors $\sum_{w \in \mathcal{R}} |\mathcal{B}_w|^2$ and the summation of the squared numbers of regular neighbors $\sum_{w \in \mathcal{R}} |\mathcal{R}_w|^2$ are close, or the step size $\alpha$ is sufficiently small.
		\end{remark}

%By omitting the constants $\eta$,  $\epsilon $ and $\alpha$ which remain unchanged across various Byzantine attacks, the learning error in Theorem 1 is $O( \sum_{w \in \mathcal{R}}\lambda^2(|\mathcal{R}_w|^2 + |\mathcal{B}_w|^2) p)$. Compare the theoretical results in Theorem 2 and Theorem 1,  and consider the  normal case that the number of Byzantine neighbors is in the same order of that of regular neighbors for each regular agent, which leads to $|\mathcal{B}_w| = O(|\mathcal{R}_w|)$ for any $w \in \mathcal{R}$, the upper bound of convergence error $\Delta$ matches the lower bound in \eqref{eq-lower-bound}. This implies that our proposed methods achieve the optimal learning error. It is worth noting that according to Theorem 2 in \cite{peng2021byzantine}, DRSA can also achieve the lower bound \eqref{eq-lower-bound} with a diminishing step size, albeit with a sublinear convergence speed. Therefore, our proposed methods not only achieve optimal convergence error but also exhibit faster convergence speed compared to DRSA.

		To further illustrate the benefits of variance reduction to Byzantine-robust decentralized stochastic optimization, we compare the theoretical results with that of DRSA.
		\begin{lemma}\label{thm3}
			\cite[Constant Step Size Variant of Theorem 2]{peng2021byzantine} Suppose Assumptions \ref{assump:network connectivity}, \ref{assump:random sampling}, \ref{assump:strong convexity}, \ref{assump:lipschitz continuous gradients}, \ref{assump:bounded variance} hold true. Denote $\eta = \min\{\frac{2\mu_w L_w}{\mu_w + L_w}\} - \epsilon > 0$ and $\epsilon \in (0, \min_{w \in \mathcal{R}}\{\frac{2\mu_w L_w}{\mu_w + L_w}\})$. Set the step size of DRSA as ${\alpha} = \min_{w \in \mathcal{R}} \left\{\frac{1}{4 (\mu_w + L_w)} \right\}.$
			We have
			\begin{align}\label{th3-1}
				\E \|x^k - x^*\|^2  \leq  (1 - \eta{\alpha})^k \|x^0 - x^*\|^2 + \Delta  + \frac{2\alpha}{\eta} \sum_{w \in \mathcal{R}} \delta_w^2,
			\end{align}
			where $\Delta$ is defined in \eqref{th2-4}.
			%	%
			%	\begin{align}\label{th3-2}
				%		{\Delta}_2 := \frac{{\alpha}}{\eta} \sum_{w \in \mathcal{R}} (32\lambda^2|\mathcal{R}_w|^2p & + 4\lambda^2|\mathcal{B}_w|^2p + 2\delta_w^2) \\
				%		&+ \frac{1}{\epsilon\eta}\sum_{w \in \mathcal{R}} \lambda^2|\mathcal{B}_w|^2p. \nonumber
				%	\end{align}
			%	%
			Here the expectation is taken over all random variables.
		\end{lemma}

		Lemma \ref{thm3} shows that with a constant step size, DRSA can also achieve a linear convergence rate, and the learning error is relative to the summed variance $\sum_{w \in \mathcal{R}}\delta_w^2$. Comparing the learning errors in Theorem \ref{thm2} and Lemma \ref{thm3}, we can observe that in BRAVO-SAGA and BRAVO-LSVRG the effect of stochastic gradient noise has been fully eliminated. On the other hand, Lemma \ref{thm3} also implies that with a diminishing step size, DRSA has a learning error that matches the lower bound in \eqref{eq-lower-bound}, but the convergence rate is sublinear. To the best of our knowledge, the proposed BRAVO-SAGA and BRAVO-LSVRG are the first Byzantine-robust decentralized stochastic optimization methods that achieve stochastic gradient noise-independent learning errors and linear convergence speeds simultaneously.

%		\red{		
%			\begin{remark}
%				It is important to note that the coefficients $\eta$ and $\epsilon$ in Theorem 1 are influenced by the strong convexity constant $\mu_w$ and the Lipschitz constant $L_w$ associated with each regular agent's local cost function. Without loss of generality, consider the case where each regular agent's cost function shares the same strong convexity constant denoted as $\mu$ and Lipschitz constant denoted as $L$, the values of $\eta$ and $\epsilon$ become related to $\frac{\mu L}{\mu + L}$. It is worth mentioning that this value lies between $\frac{\mu}{2}$ and $\frac{L}{2}$. Consequently, if the cost function of each regular agent exhibits more favorable optimization properties, such as a larger strong convexity constant $\mu$, we can choose larger values for $\eta$ and $\epsilon$. This choice, as described in Theorem 1, leads to faster convergence speeds and smaller convergence errors for BRAVO-SAGA and BRAVO-LSVRG.
%			\end{remark}
%		}
		
		\section{Numerical Experiments}
		In this section, we conduct extensive numerical experiments to demonstrate the Byzantine-robustness of the proposed methods, BRAVO-SAGA and BRAVO-LSVRG.
		
		We consider an undirected Erdos-Renyi graph consisting of $N = 100$ agents, in which every edge $e = (w, v)$ for any $w, v \in \mathcal{V}$ is connected with probability $q \in [0, 1]$. We set $q = 0.5$. All experiments are conducted on the MNIST and Fashion-MNIST datasets using softmax regression. The global cost function is defined as
		\begin{align}
			F(\tilde{x}) = - \frac{1}{J}\sum_{j=1}^J\sum_{m=0}^{M-1} \left(I(b^{(j)} = m) \ln \left(\frac{\exp((\tilde{x})_m^T a^{(j)})}{\sum_{l=0}^{M-1} \exp((\tilde{x})_l^T a^{(j)})}\right)\right). \nonumber
		\end{align}
		Here $J$ and $M$ are the numbers of samples and categories, respectively. Sample $j$ is represented by $(a^{(j)}, b^{(j)})$, where $a^{(j)} \in \R^{p/M}$ is the data and $b^{(j)} \in \R$ is the target. $I(b^{(j)} = m)$ is the indicator function; $I(b^{(j)} = m) = 1$ if $b^{(j)} = m$, and $I(b^{(j)} = m) = 0$ otherwise. The model is $\tilde{x} \in \R^p$ and $(\tilde{x})_m \in \R^{p / M}$ is the $m$-th block of $\tilde{x}$. The MNIST dataset contains $M = 10$ handwritten digits from 0 to 9, with 60,000 training images and 10,000 testing images whose dimensions are $p/M=784$. The Fashion-MNIST dataset contains fashion images and the other attributes are the same as those of the MNIST dataset.
		
		%\red{\st{The Fashion-MNIST dataset is only used in demonstrating the optimality gap. }}
		
		%The Fashion-MNIST dataset only serves for experiments on optimality gap.
		
		We consider i.i.d. (independent and identically distributed) and non-i.i.d. data distributions across the agents. In the i.i.d. case, all $N$ agents randomly and evenly partition the training samples. In the non-i.i.d. case, every 10 agents randomly and evenly partition the training samples of one class.
		
		%\subsection{Benchmark methods}
		We compare our proposed methods BRAVO-SAGA and BRAVO-LSVRG with several benchmark methods, including DPSGD \cite{lian2017can,nedic2009distributed}, the stochastic version of ByRDiE \cite{yang2019byrdie} (called ByRDiE-S), the stochastic version of BRIDGE \cite{yang2019bridge} (called BRIDGE-S), and DRSA \cite{peng2021byzantine}. For DPSGD, we choose the Metropolis weights \cite{boyd2004distributed} to ensure the mixing matrix is doubly stochastic.
		For ByRDiE-S, we set the number of inner iteration to be 1, and for fair comparison, refer one iteration as all dimensions being updated once. All step sizes of the benchmark methods are hand-tuned to the best. For DRSA, the step size is set to $\alpha^k = O(\frac{1}{\sqrt{k}})$ as in \cite{peng2021byzantine}, unless otherwise specified.
		
		The performance metrics include classification accuracy, model variance, and convergence error $\|x^k-x^*\|^2$. When computing the classification accuracy over the testing samples, we randomly choose one regular agent to compute the classification accuracy on its local model. To see the consensus error between the regular agents, we also show the variance of the regular models. See \url{http://github.com/pengj97/BRAVO}.
		
		%
%				\begin{figure}
%%			\subfigure[MNIST dataset]{
%%				\includegraphics[width=0.47\linewidth]{figure/acc-wa.pdf}
%%				\includegraphics[width=0.47\linewidth]{figure/var-wa.pdf}
%%				\label{fig:wa-a}}
%			%	\subfigure[Fashion-MNIST dataset]{
%				%		\includegraphics[width=0.47\linewidth]{figure/fmnist-acc-wa.pdf}
%				%		\includegraphics[width=0.47\linewidth]{figure/fmnist-var-wa.pdf}
%				%		\label{fig:wa-b}}
%			\includegraphics[scale=0.28]{figure/wa.pdf}
%			\caption{Classification accuracy and variance of agents' local models without Byzantine attacks with i.i.d. data.}
%			\label{fig:wa}
%		\end{figure}
		
%		\begin{figure}
%			\begin{minipage}{0.49\linewidth}
%				\centering
%				\includegraphics[width=8.2cm]{figure/wa.pdf}
%				\caption{Classification accuracy and variance of agents' local models without Byzantine attacks with i.i.d. data.}
%				\label{fig:wa}
%			\end{minipage}
%			\hfill
%			\begin{minipage}{0.49\linewidth}
%				\centering
%				\includegraphics[width=8.2cm]{figure/ga.pdf}
%				\caption{Classification accuracy and variance of regular agents' local models under Gaussian attacks with i.i.d. data.}				
%				\label{fig:ga}
%			\end{minipage}
%		\end{figure}
%%		

		%
		\begin{figure}[t]
			\centering
			\hspace{-32pt}
			\includegraphics[scale=0.22]{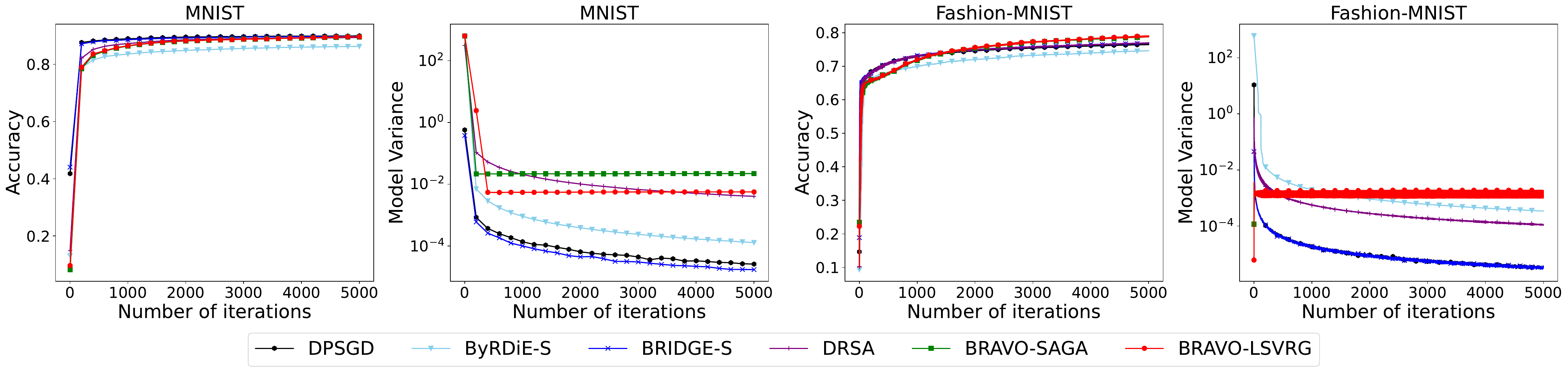}
			\caption{Classification accuracy and variance of agents' local models without Byzantine attacks with i.i.d. data.}
			\label{fig:wa}
		\end{figure}
		\begin{figure}[t]
			\centering
			\hspace{-32pt}
			\includegraphics[scale=0.22]{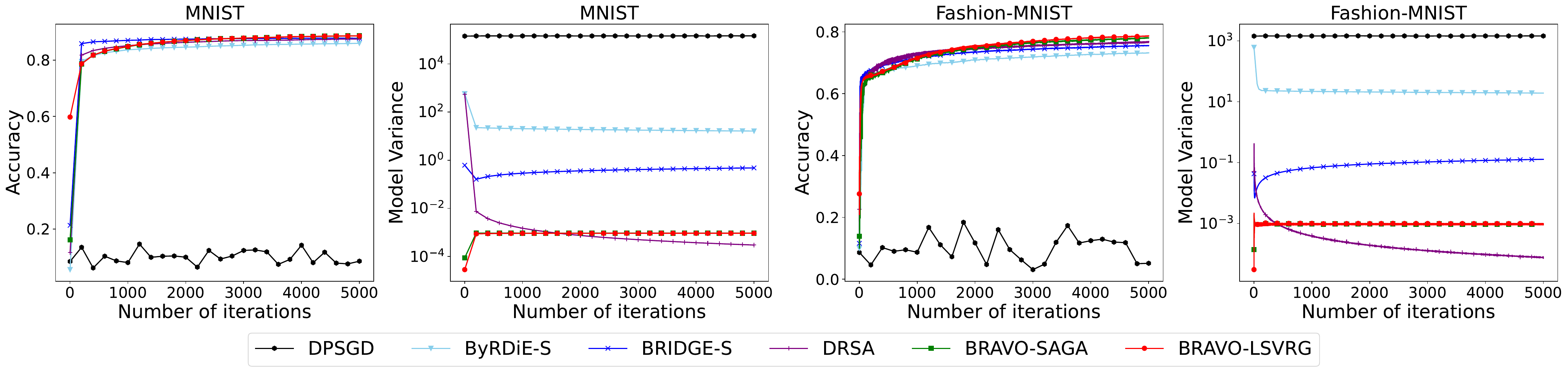}
				\caption{Classification accuracy and variance of regular agents' local models under Gaussian attacks with i.i.d. data.}				
				\label{fig:ga}
		\end{figure}

		\subsection{Byzantine attacks}
		When Byzantine agents exist, we set their number as $B = 20$, unless otherwise specified. We randomly choose $B$ agents to be Byzantine but guarantee the network of regular agents to be connected. We consider the following Byzantine attacks in the numerical experiments.
		
		\noindent\textbf{Gaussian attacks.} Every Byzantine agent $v \in \mathcal{B}$ sends to its neighbors malicious model $x_v^k$, whose entries independently follow the Gaussian distribution $\mathcal{N}(0, 100^2)$.
		
		\noindent\textbf{Sign-flipping attacks.} Every Byzantine agent $v \in \mathcal{B}$ calculates its true model $x_v^k$, multiplies with a negative constant $c$, and sends the malicious model $z_v^k = c x_v^k$ to its neighbors. Here we set $c = -4$.
		
		%	\noindent\textbf{Same-value attacks} Every Byzantine agent $j \in \mathcal{B}$ sends the malicious model $z_j^k = c * \mathbf{1}$ to its all neighbors where $\mathbf{1}$ is an all-one vector and $c$ is a constant. We set $c=100$ here.
		
		%	\noindent\textbf{Gaussian attacks} Every Byzantine agent $j \in \mathcal{B}$ sends the malicious model $z_j^k$, whose elements follow Gaussian distribution $\mathcal{N}(0, 10000)$, to its all neighbors.
		
		%	\noindent\textcolor{red}{\textbf{ALIE attacks [ALIE]} Every Byzantine agent $j \in \mathcal{B}$ estimates the mean $\mu_d$ and variance $\sigma_d$ of every dimension $d \in [p]$ of regular models and sends $[(z_j^k)_d = \mu_d + z^{max} \cdot \sigma_d]_{d \in [p]}$ to its all neighbors where $z^{max} = \max_z \left(\phi(z) < \frac{n - b}{n} \right)$ and we get $z^{max} = 0.84$. }
		
		\noindent\textbf{Sample-duplicating attacks.} The Byzantine agents collude to select a specific regular agent, copy the model of the selected agent, and send it to neighbors. This amounts to that the Byzantine agents duplicate the samples of the selected regular agent. Here the sample-duplicating attacks are only applied to the non-i.i.d. case.

%		%
%		\begin{figure}
%			\begin{minipage}{0.49\linewidth}
%				\centering
%				\includegraphics[width=8.2cm]{figure/sf.pdf}
%				\caption{Classification accuracy and variance of regular agents' local models under sign-flipping attacks with i.i.d. data.}
%				\label{fig:sf}
%			\end{minipage}
%			\hfill
%			\begin{minipage}{0.49\linewidth}
%				\centering
%				\includegraphics[width=8.2cm]{figure/inner-var-wa.pdf}
%				\caption{Summed variance of regular agents' (corrected) stochastic gradients without Byzantine attacks with i.i.d. data.}
%				\label{fig:inner}
%			\end{minipage}
%		\end{figure}
%		

			%
	\begin{figure}[t]
		\centering
		\hspace{-32pt}
		\includegraphics[scale=0.22]{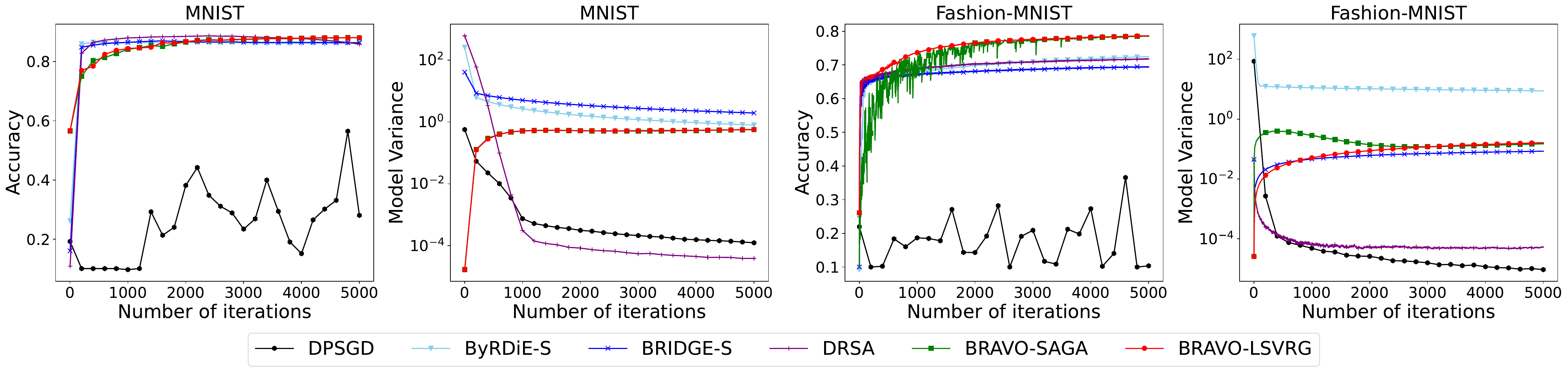}
				\caption{Classification accuracy and variance of regular agents' local models under sign-flipping attacks with i.i.d. data.}
				\label{fig:sf}
	\end{figure}
	\begin{figure}
					\centering
					\includegraphics[scale=0.22]{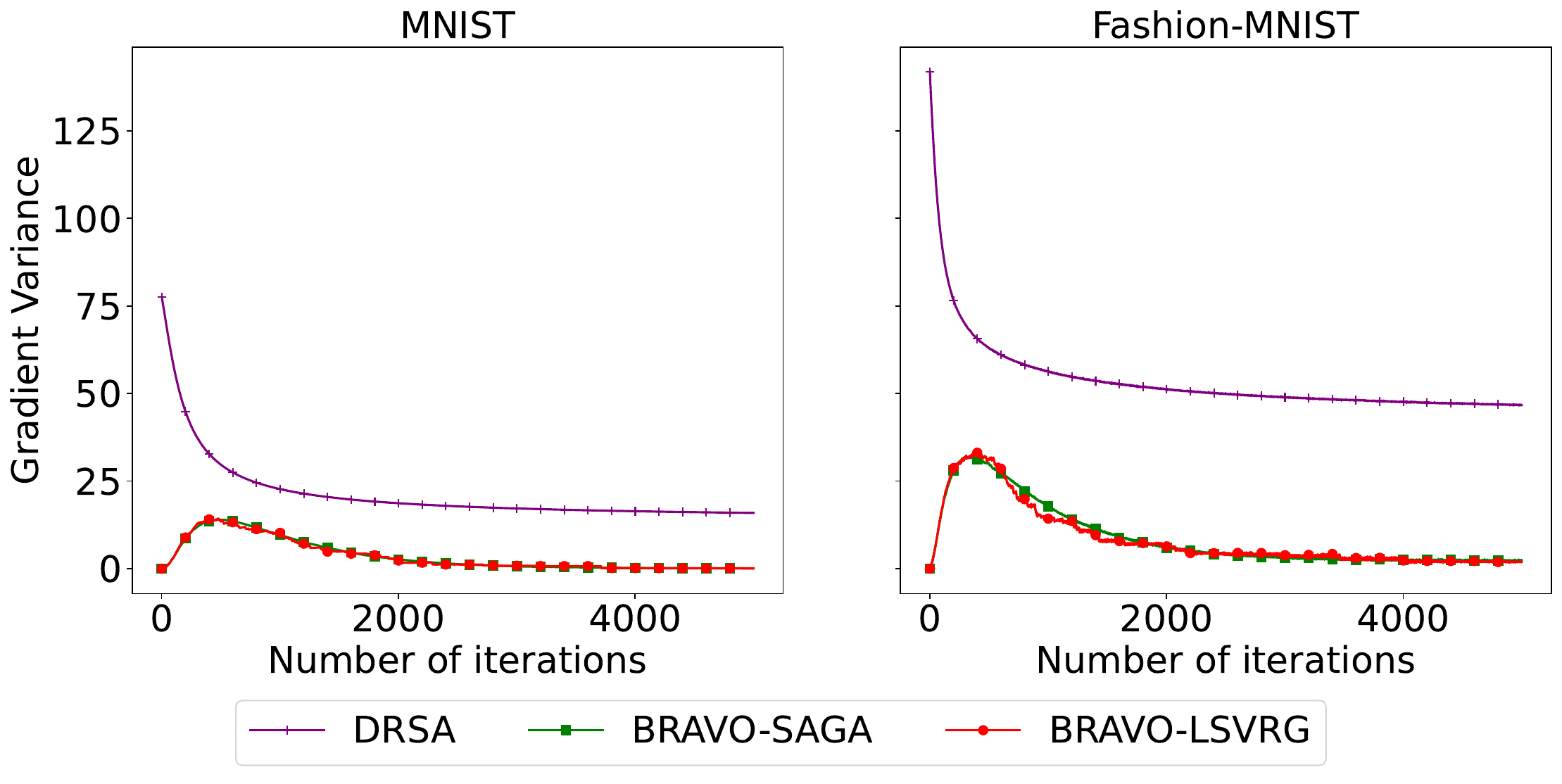}
					\caption{Summed variance of regular agents' (corrected) stochastic gradients without Byzantine attacks with i.i.d. data.}
					\label{fig:inner}

		\end{figure}
	%
		%
%		\begin{figure}
%			%	\subfigure[MNIST ]{
%				%		\includegraphics[width=0.47\linewidth]{figure/acc-sf.pdf}
%				%		\includegraphics[width=0.47\linewidth]{figure/var-sf.pdf}
%				%		\label{fig:sf-a}}
%			%	\subfigure[Fashion-MNIST]{
%				%		\includegraphics[width=0.47\linewidth]{figure/fmnist-acc-sf.pdf}
%				%		\includegraphics[width=0.47\linewidth]{figure/fmnist-var-sf.pdf}
%				%		\label{fig:sf-b}}
%			\includegraphics[scale=0.28]{figure/sf.pdf}
%			\caption{Classification accuracy and variance of regular agents' local models under sign-flipping attacks with i.i.d. data.}
%			\label{fig:sf}
%		\end{figure}
%		%
%		
%		\begin{figure}
%			\centering
%			\includegraphics[width=0.90\linewidth]{figure/inner-var-wa.pdf}
%			\caption{Summed variance of regular agents' (corrected) stochastic gradients without Byzantine attacks with i.i.d. data.}
%			\label{fig:inner}
%		\end{figure}

		\subsection{Experiments on the i.i.d. case}
		
		\noindent\textbf{Without attacks.} When the data distribution is i.i.d. and the number of Byzantine agents is $B=0$, the numerical results are shown in Fig. \ref{fig:wa}. In BRAVO-SAGA and BRAVO-LSVRG, the penalty parameter is $\lambda = 0.005$ and the step size is $\alpha = 0.01$. All methods perform well in this case, although ByRDiE-S is slower than the others. On the Fashion-MNIST dataset, BRAVO-SAGA and BRAVO-LSVRG reach slightly higher accuracy than DRSA.
		
		\noindent\textbf{Gaussian attacks and sign-flipping attacks.} Set $\lambda = 0.005$ and $\alpha = 0.01$ under Gaussian attacks, as well as $\lambda = 0.0001$ and $\alpha = 0.01$ under sign-flipping attacks for BRAVO-SAGA and BRAVO-LSVRG. The numerical results are shown in Figs. \ref{fig:ga} and \ref{fig:sf}. DPSGD fails since it is not designed to defend against Byzantine attacks. All Byzantine-robust methods perform well, while BRAVO-SAGA and BRAVO-LSVRG demonstrate faster convergence speeds and better classification accuracies, especially in the Fashion-MNIST dataset.

		\noindent\textbf{Importance of variance reduction.} As we can observe from Figs. \ref{fig:ga} and \ref{fig:sf}, the performance gains of both BRAVO-SAGA and BRAVO-LSVRG over DRSA are more obvious in the Fashion-MNIST dataset than in MNIST. To investigate this phenomenon, we depict how the summed variance of the regular agents' (corrected) stochastic gradients evolves for the three methods and on the two datasets, as shown in Fig. \ref{fig:inner}. We consider that there is no Byzantine agent so as to avoid the possible interference of Byzantine attacks. The parameters are the same as those in plotting Fig. \ref{fig:wa}. Observe that with the variance reduction methods, the stochastic gradient noise is indeed reduced. Also observe that the level of the uncorrected stochastic gradient noise on MNIST is much lower than that on Fashion-MNIST. Hence, the improvement of BRAVO-SAGA and BRAVO-LSVRG over DRSA is smaller. The numerical results corroborate the theoretical findings that the learning error of DRSA is relative to the summed variance $\sum_{w \in \mathcal{R}}\delta_w^2$, while the learning errors of the two variance-reduced ones are independent with the stochastic gradient noise.
		
		% \red{whereas DECEMBER-SAGA and DECEMBER-LSVRG have faster convergence rate and achieve smaller learning error than all other methods as shown in Fig. \ref{fig:ga-b}. }
		%
		%\red{, except that the performance gain of DECEMBER-SAGA and DECEMBER-LSVRG is more obvious, which is depicted in Fig. \ref{fig:sf-b}.}

%		\begin{figure}
%			\begin{minipage}{0.49\linewidth}
%				\centering
%					\includegraphics[width=8.2cm]{figure/sd.pdf}
%				\caption{Classification accuracy and variance of regular agents' local models under sample-duplicating attacks with non-i.i.d. data.}
%				\label{fig:sd}
%			\end{minipage}
%			\hfill
%			\begin{minipage}{0.49\linewidth}
%				\centering
%				\includegraphics[width=8.2cm]{figure/mix.pdf}
%				\caption{Classification accuracy and variance of regular agents' local models under mixture type of Byzantine attacks (where half of Byzantine agents use Gaussian attacks and another half use sample-duplicating attacks) with non-i.i.d. data.}
%				\label{fig:mix}
%			\end{minipage}
%		\end{figure}
%		
				%
		\begin{figure}[t]
			\centering
			\hspace{-32pt}
			\includegraphics[scale=0.22]{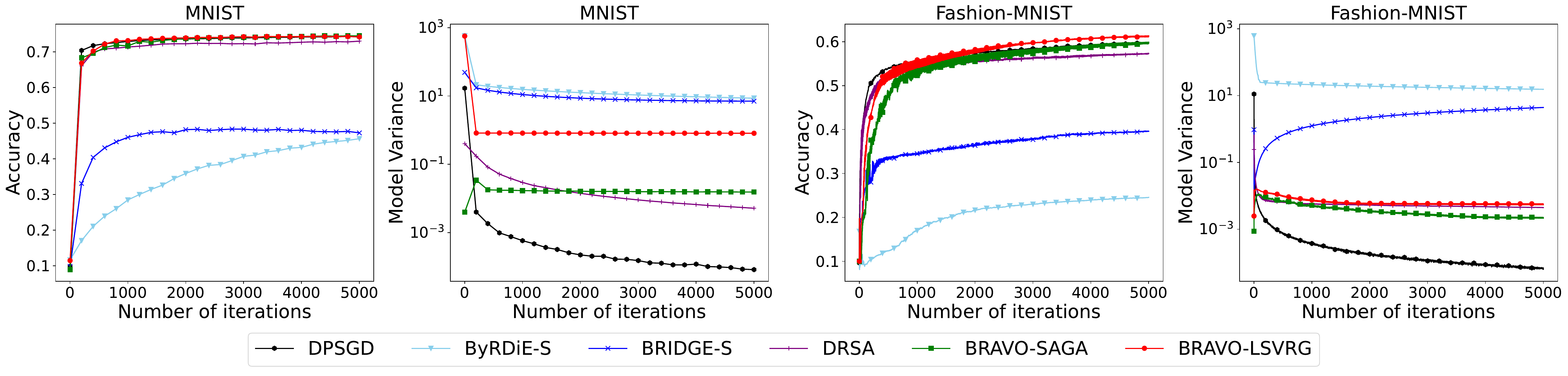}
				\caption{Classification accuracy and variance of regular agents' local models under sample-duplicating attacks with non-i.i.d. data.}
				\label{fig:sd}
		\end{figure}
		\begin{figure}[t]
			\centering
			\hspace{-32pt}
			\includegraphics[scale=0.22]{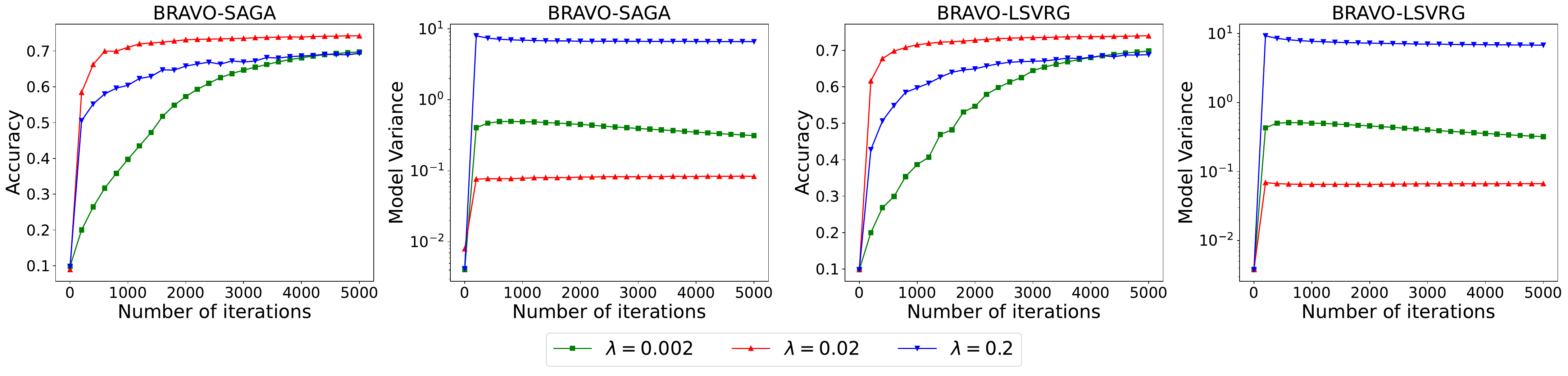}
			\caption{Impact of penalty parameter $\lambda$ for BRAVO-SAGA and BRAVO-LSVRG under sample-duplicating attacks with non-i.i.d. data. }
			\label{fig:imopp}
		\end{figure}
		\begin{figure}
							\centering
			\includegraphics[scale=0.22]{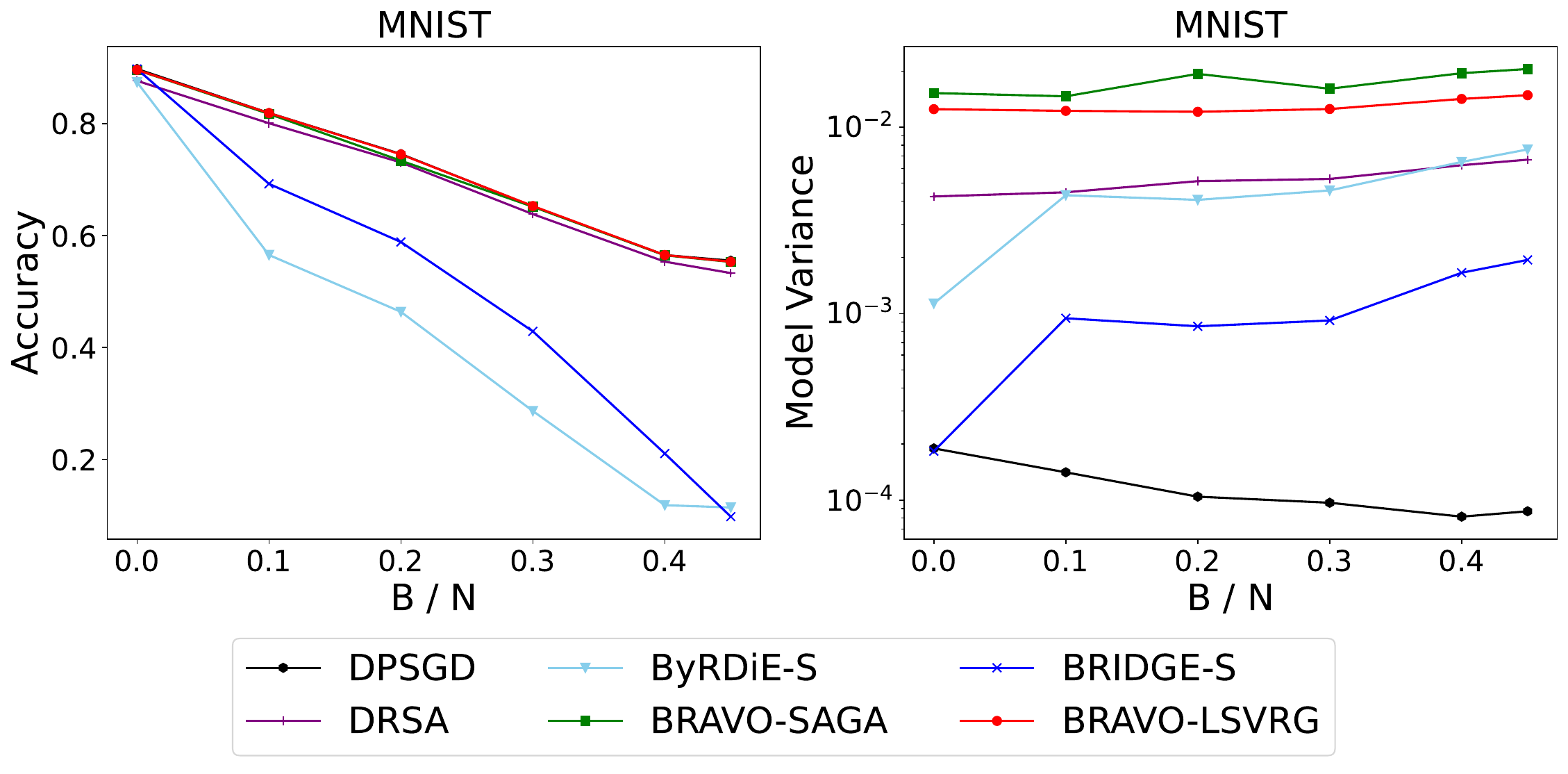}
			\caption{Classification accuracy of regular agent's local models with different fractions of Byzantine agents under sample-duplicating attacks with non-i.i.d. data.}
			\label{fig:byz_frac}
		\end{figure}
		
		%
%		\begin{figure}[htbp]
%			\centering
%			%	\hspace{-18pt}
%			%	\subfigure[DECEMBER-SAGA]{
%				%		\centering
%				%		\includegraphics[width=0.47\linewidth]{figure/imopp-acc-saga.pdf}
%				%		%\hspace{5pt}
%				%		\includegraphics[width=0.47\linewidth]{figure/imopp-var-saga.pdf}
%				%		\hspace{-17pt}
%				%	}
%			%
%			%	\hspace{-18pt}
%			%	\subfigure[DECEMBER-LSVRG]{
%				%		\centering
%				%		\includegraphics[width=0.47\linewidth]{figure/imopp-acc-lsvrg.pdf}
%				%		%\hspace{5pt}
%				%		\includegraphics[width=0.47\linewidth]{figure/imopp-var-lsvrg.pdf}
%				%		\hspace{-17pt}
%				%	}
%			
%			%\vspace{-10pt}
%			\includegraphics[scale=0.28]{figure/imopp.pdf}
%			\caption{Impact of penalty parameter $\lambda$ for BRAVO-SAGA and BRAVO-LSVRG under sample-duplicating attacks with non-i.i.d. data. }
%			\label{fig:imopp}
%		\end{figure}
%		%
%		%
%		\begin{figure}
%			\centering
%			%	\includegraphics[width=0.47\linewidth]{figure/acc-byz-frac.pdf}
%			%	\includegraphics[width=0.47\linewidth]{figure/var-byz-frac.pdf}
%			\includegraphics[scale=0.28]{figure/byz-frac.pdf}
%			\caption{Classification accuracy of regular agent's local models with different fractions of Byzantine agents under sample-duplicating attacks with non-i.i.d. data.}
%			\label{fig:byz_frac}
%			
%		\end{figure}
		%
		
		\subsection{Experiments on the non-i.i.d. case}
		
		\noindent\textbf{Sample-duplicating attacks.} Consider the case that the data distribution is non-i.i.d. and the Byzantine attacks are sample-duplicating. It is worth noting that since the Byzantine agents own the samples of two classes, the best accuracy that the regular agents can obtain is 0.8. In BRAVO-SAGA and BRAVO-LSVRG, the penalty parameter is $\lambda = 0.02$ and the step size is $\alpha = 0.01$. As depicted in Fig. \ref{fig:sd}, BRAVO-SAGA, BRAVO-LSVRG and DRSA all perform well and achieve near-optimal accuracy. The TV-norm penalty forces the regular models to reach consensus, and thus these methods are insensitive to the non-i.i.d. data distribution. In contrast, with the majority-voting scheme, ByRDiE-S and BRIDGE-S fail in this case.
		
%		\red{\noindent\textbf{Mixture type of Byzantine attacks.} In order to further evaluate the robustness of BRAVO-SAGA and BRAVO-LSVRG, we introduce a more complex scenario where different Byzantine agents employ distinct types of Byzantine attacks. Specifically, we divide the Byzantine agents into two groups: one group utilizes Gaussian attacks, while the other group utilizes sample-duplicating attacks. In BRAVO-SAGA and BRAVO-LSVRG, the penalty parameter is $\lambda = 0.011$ and the step size is $\alpha = 0.01$. As illustrated in Fig. \ref{fig:mix}, both BRAVO-SAGA and BRAVO-LSVRG exhibit robust performance even in the presence of this mixture type of Byzantine attacks.}
		
		%\red{Figure \ref{fig:sd-b} shows that DECEMBER-SAGA and DECEMBER-LSVRG converge faster than DECEMBER obviously, which matches our theoretical results in Section \ref{sec:Theoretical Analysis} and verifies that the stochastic gradient noise plays a critical role in Byzantine-robust optimization with non-i.i.d. data.  }
		
		\subsection{Impact of $\lambda$ and $B/N$}
		
		\noindent\textbf{Impact of penalty parameter $\lambda$.} We investigate the impact of the penalty parameter $\lambda$. The settings are same as those in sample-duplicating attacks. As shown in Fig. \ref{fig:imopp}, if we choose a relatively large or small value of $\lambda$, such as $\lambda = 0.2$ or $\lambda = 0.002$, the performance of BRAVO-SAGA and BRAVO-LSVRG is not as good as that with a proper value of $\lambda$. For example, here $\lambda=0.02$ gives a favorable trade-off between convergence speed and learning error.
		
		\noindent\textbf{Impact of fraction of Byzantine agents $B/N$.} We consider the impact of the fraction of the Byzantine agents. Here the settings are the same as those in sample-duplicating attacks. Considering that ByRDiE-S and BRIDGE-S require $N > 2B$, we let the fraction of Byzantine agents be less than 0.5. In BRAVO-SAGA and BRAVO-LSVRG, the penalty parameter is $\lambda = 0.005$ and the step size is $\alpha = 0.01$.  As depicted in Fig. \ref{fig:byz_frac}, with the increasing fraction of Byzantine agents, all methods suffer from performance losses, but the advantages of BRAVO-SAGA and BRAVO-LSVRG are obvious compared to ByRDiE-S and BRIDGE-S.
		
		%The experiment results show that DECEMBER-SAGA and DECEMBER-LSVRG have good robustness under different fractions of Byzantine agents and non-i.i.d. data and better than other decentralized Byzantine-robust methods.

		\subsection{Experiments on convergence error}
		%%
		%\begin{figure}[htbp]
			%	\centering
			%	
			%	\subfigure[Without attacks]{
				%		\centering
				%		\includegraphics[width=1.6in]{figure/para-wa-iid.pdf}
				%	}
			%	\subfigure[Gaussian attacks]{
				%		\centering
				%		\includegraphics[width=1.6in]{figure/para-ga-iid.pdf}
				%	}
			%	
			%	\subfigure[Sign-flipping attacks]{
				%		\centering
				%		\includegraphics[width=1.6in]{figure/para-sf-iid.pdf}
				%	}
			%	\subfigure[Sample-duplicating attacks]{
				%		\centering
				%		\includegraphics[width=1.6in]{figure/para-sd-noniid.pdf}
				%	}
			%	
			%	%\vspace{-10pt}
			%	\caption{Optimalily gaps of DECEMBER ($\alpha^k = 0.01 / k$), DECEMBER ($0.01 / \sqrt{k}$), DECEMBER-SAGA ($\alpha = 0.01$) and DECEMBER-LSVRG ($\alpha = 0.01$) on MNIST dataset. }
			%	\label{fig:opt-gap-mnist}
			%\end{figure}
		%%
		%%
		%\begin{figure}[htbp]
			%	\centering
			%	\subfigure[Without attacks]{
				%		\centering
				%		\includegraphics[width=1.6in]{figure/fashion-para-wa-iid.pdf}
				%	}
			%	\subfigure[Gaussian attacks]{
				%		\centering
				%		\includegraphics[width=1.6in]{figure/fashion-para-ga-iid.pdf}
				%	}
			%	
			%	\subfigure[Sign-flipping attacks]{
				%		\centering
				%		\includegraphics[width=1.6in]{figure/fashion-para-sf-iid.pdf}
				%	}
			%	\subfigure[Sample-duplicating attacks]{
				%		\centering
				%		\includegraphics[width=1.6in]{figure/fashion-para-sd-noniid.pdf}
				%	}
			%	
			%	%\vspace{-10pt}
			%	\caption{Optimalily gaps of DECEMBER ($\alpha^k = 0.01 / k$), DECEMBER ($0.01 / \sqrt{k}$), DECEMBER-SAGA ($\alpha = 0.01$) and DECEMBER-LSVRG ($\alpha = 0.01$) on Fashion-MNIST dataset. }
			%	\label{fig:opt-gap-fashion}
			%\end{figure}
		%%
		
				%
		\begin{figure}[t]
			\centering
			\hspace{-32pt}
			\includegraphics[scale=0.22]{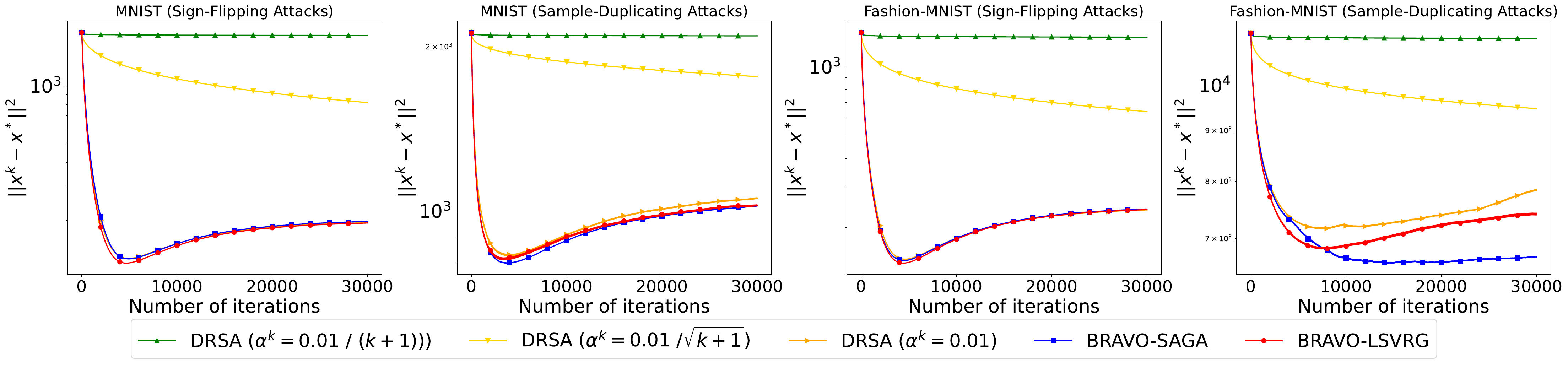}
			\caption{Convergence error of DRSA ($\alpha^k = 0.01 / (k+1)$), DRSA ($0.01 / \sqrt{k+1}$), DRSA ($\alpha = 0.01$), BRAVO-SAGA ($\alpha = 0.01$) and BRAVO-LSVRG ($\alpha = 0.01$) under sign-flipping attacks with i.i.d. data and sample-duplicating attacks with non-i.i.d. data.}
			\label{fig:opt-gap}
		\end{figure}
		%
		
%		%
%		\begin{figure}[t]
%			\centering
%			\includegraphics[width=8.2cm]{figure/optimality-gap.pdf}
%			\caption{Convergence error of DRSA ($\alpha^k = 0.01 / (k+1)$), DRSA ($0.01 / \sqrt{k+1}$), DRSA ($\alpha = 0.01$), BRAVO-SAGA ($\alpha = 0.01$) and BRAVO-LSVRG ($\alpha = 0.01$) under sign-flipping attacks with i.i.d. data and sample-duplicating attacks with non-i.i.d. data.}
%			\label{fig:opt-gap}
%		\end{figure}
%		%
		
		To further highlight the advantages of BRAVO-SAGA and BRAVO-LSVRG over DRSA, we also compare them on the MNIST and Fashion-MNIST datasets in terms of the convergence error $\|x^k-x^*\|^2$. Since the step sizes of DRSA are different in theory and in practice, we choose three step size rules for DRSA, $\alpha^k = 0.01/(k+1)$, $\alpha^k = 0.01/\sqrt{k+1}$ and $\alpha^k = 0.01$. The attacks are sign-flipping with i.i.d. data where we set $\lambda = 0.0001$ and sample-duplicating with non-i.i.d. data where we set $\lambda = 0.02$. The experimental results are depicted in Fig. \ref{fig:opt-gap}. Observe that BRAVO-LSVRG and BRAVO-SAGA, with the help of variance reduction, both achieve faster convergence speeds than DRSA with the diminishing step sizes, and achieve smaller learning errors than DRSA with a constant step size that suffers from the stochastic gradient noise. These experimental results corroborate the theoretical analysis in Section \ref{sec:Theoretical Analysis} and show the superior performance of BRAVO-SAGA and BRAVO-LSVRG.
		
		\section{Conclusions}
		
		This paper aims at developing Byzantine-robust stochastic optimization methods over a decentralized network, in which an unknown number of Byzantine agents collude to bias the optimization process. Motivated by fact that the stochastic gradient noise significantly affects the learning error, we develop two variance-reduced Byzantine-robust decentralized stochastic optimization methods, BRAVO-SAGA and BRAVO-LSVRG. These methods enjoy linear convergence speeds and provable learning errors, which are independent of the stochastic gradient noise and are shown to be optimal in order for any subgradient-based method solving the TV-norm penalized decentralized stochastic optimization problem.
		
		In light of this work, one natural question arises: What is the lower bound of the learning errors of first-order methods when applied to solving the non-penalized  Byzantine-robust decentralized stochastic optimization problem \eqref{problem-original}? Answering this fundamental question is challenging in our current setting, since we impose no restrictions on the data distributions across the agents. We will investigate this issue for the i.i.d. data distribution in our future work.
		
		\vspace{1em}
		\noindent \textbf{Acknowledgement.} Qing Ling is supported in part by NSF China Grants 61973324 and 12126610, Guangdong Basic and Applied Basic Research Foundation Grant 2021B1515020094, and Guangdong Provincial Key Laboratory of Computational Science Grant 2020- B1212060032. A short preliminary version of this paper has appeared in IEEE International Conference on Acoustics, Speech and Signal Processing, Singapore, May 22--27, 2022 \cite{peng2022variance}.
		\vspace{1em}
		
		%\blue{In this paper, we propose two  Byzantine-robust decentralized finite-sum optimization methods. Theoretical results and experimental results show the linear convergence rate and robustness of our proposed methods. With the help of variance reduction, our proposed methods gradually eliminate the variance of stochastic gradients and achieve smaller learning errors, which is independent with the stochastic gradient noise.}
		
		%\section*{Acknowledgment}


		%\newpage

		% or
		

		\appendix
		%	\vspace{1em}
		
		%\newpage
		%\onecolumn
		\section{Proof of Theorem 1}

		\begin{proof}
			We prove Theorem 1 for BRAVO-LSVRG first, and then modify the proof for BRAVO-SAGA.
			%We first take conditional expectation given the variables up to time $k$, namely $\{x_w^l: l \leq k, w \in \mathcal{R}\}$, and take the expectation over $\{x_w^l, l \leq k, w \in \mathcal{R}\}$. For simplicity, we denote $\E[\cdot | x_w^l: l \leq k, w \in \mathcal{R}]$ as $\E_k[\cdot]$.
			
			\textbf{Step 1.}  We begin with taking the conditional expectation given the random variables up to time $k$, namely $\{i_w^l: l < k, w \in \mathcal{R}\}$, and then taking the expectation over these random variables. For notational simplicity, denote the conditional expectation $\E[\cdot|i_w^l: l < k, w \in \mathcal{R}]$ as $\E_k[\cdot]$. From the update in \eqref{eq-DECEMBER-VR}, for every regular agent $w$, we have
			\vspace{-17pt}
			\begin{align}\label{proof:eq1}
				&\E_k \|x_w^{k+1} - x_w^*\|^2 \\
				=&\E_k\|x_w^k - x_w^* - \alpha \cdot \Big(g_w^k + \lambda \sum_{v \in \mathcal{R}_w} sign(x_w^k - x_v^k)\Big) -  \alpha \cdot\Big( \lambda \sum_{v \in \mathcal{B}_w} sign(x_w^k - z_v^k)\Big)\|^2 \nonumber \\
				=&\|x_w^k - x_w^*\|^2 +\alpha^2 \cdot \E_k\|g_w^k + \lambda \sum_{v \in \mathcal{R}_w} sign(x_w^k - x_v^k)  + \lambda \sum_{v \in \mathcal{B}_w} sign(x_w^k - z_v^k)\|^2 \nonumber \\
				&-2\alpha \cdot \langle \E_k[g_w^k] + \lambda \sum_{v \in \mathcal{R}_w} sign(x_w^k - x_v^k), x_w^k - x_w^* \rangle  -2\alpha \cdot \langle\lambda \sum_{v \in \mathcal{B}_w} sign(x_w^k - z_v^k), x_w^k - x_w^* \rangle. \nonumber
			\end{align}

				\vspace{-17pt}
			We handle the terms at the right-hand side of \eqref{proof:eq1} one by one.
			
			For the second term at the right-hand side of \eqref{proof:eq1}, consider the optimality condition of \eqref{problem-TV}, given by
			\vspace{-17pt}
			\begin{align}\label{proof:eq2}
				F'_w(x_w^*) + \lambda \sum_{v \in \mathcal{R}_w} s_{wv} = 0,
			\end{align}

			\vspace{-17pt}
			\noindent which holds true for some $s_{wv} \in [-1, 1]^p$ and $s_{vw} = - s_{wv}$. Then we have
			\vspace{-17pt}
			\begin{align}\label{proof:eq3}
				&\E_k\|g_w^k + \lambda \sum_{v \in \mathcal{R}_w} sign(x_w^k - x_v^k) + \lambda \sum_{v \in \mathcal{B}_w} sign(x_w^k - z_v^k)\|^2 \nonumber \\
				=&\E_k\|F'_w(x_w^k) + \lambda \sum_{v \in \mathcal{R}_w} sign(x_w^k - x_v^k) + \lambda \sum_{v \in \mathcal{B}_w} sign(x_w^k - z_v^k)- F'_w(x_w^*) - \lambda \sum_{v \in \mathcal{R}_w} s_{wv} + g_w^k - F'_w(x_w^k)\|^2 \nonumber\\
				\leq& 2\|F'_w(x_w^k) + \lambda \sum_{v \in \mathcal{R}_w} sign(x_w^k - x_v^k) + \lambda \sum_{v \in \mathcal{B}_w} sign(x_w^k - z_v^k)  \ - F'_w(x_w^*) - \lambda \sum_{v \in \mathcal{R}_w} s_{wv} \|^2+ 2\E_k \|g_w^k - F'_w(x_w^k) \|^2 \nonumber \\
				\leq&4\E_k \|F'_w(x_w^k) - F'_w(x_w^*)   + \lambda \sum_{v \in \mathcal{R}_w} sign(x_w^k - x_v^k) - \lambda \sum_{v \in \mathcal{R}_w} s_{wv} \|^2 + 4\|\lambda \sum_{v \in \mathcal{B}_w} sign(x_w^k - z_v^k)\|^2 +  2\E_k \|g_w^k - F'_w(x_w^k) \|^2 \nonumber \\
%			\end{align}
%		\vspace{-10pt}
%		\vspace{-10pt}
%			\begin{align}\label{proof:eq3}
				\leq&8\E_k \|F'_w(x_w^k) - F'_{w}(x_w^*)\|^2 + 8 \|\lambda \sum_{v \in \mathcal{R}_w} sign(x_w^k - x_v^k) - \lambda \sum_{v \in \mathcal{R}_w} s_{wv} \|^2 + 4\|\lambda \sum_{v \in \mathcal{B}_w} sign(x_w^k - z_v^k)\|^2+  2\E_k \|g_w^k - F'_w(x_w^k) \|^2 \nonumber\\
				\leq&8\E_k \|F'_w(x_w^k) - F'_{w}(x_w^*)\|^2 + 32 \lambda^2 |\mathcal{R}_w|^2 p + 4\lambda^2|\mathcal{B}_w|^2p  +  2\E_k \|g_w^k - F'_w(x_w^k) \|^2.
			\end{align}

			\vspace{-17pt}
			For the last term at the right-hand side of \eqref{proof:eq3}, from \eqref{eq-lsvrg}, we have
			\vspace{-17pt}
			\begin{align}\label{proof:eq4}
				&\E_k \|g_w^k - F'_w(x_w^k)\|^2 \\
				=&\E_k\|F'_{w, i_w^k}(x_w^k) - F'_{w, i_w^k} (x_w^*) + F'_{w}(x_w^*) - F'_w(x_w^k)- (F'_{w, i_w^k}(y_w^k) - F'_{w, i_w^k}(x_w^*) + F'_w(x_w^*) - F'_{w}(y_w^k) ) \|^2 \nonumber\\
				\leq&2\E_k\|F'_{w, i_w^k}(x_w^k) - F'_{w, i_w^k} (x_w^*) + F'_{w}(x_w^*) - F'_w(x_w^k)\|^2  + 2\E_k \|F'_{w, i_w^k}(y_w^k) - F'_{w, i_w^k}(x_w^*) + F'_w(x_w^*) - F'_{w}(y_w^k)\|^2 \nonumber\\
				\leq&2\E_k \|F'_{w, i_w^k}(x_w^k) - F'_{w, i_w^k} (x_w^*)\|^2 + 2\E_k \|F'_{w, i_w^k}(x_w^*) - F'_{w, i_w^k}(y_w^k)\|^2 \nonumber\\
				\leq&2L_w^2  \|x_w^k - x_w^*\|^2 + 2L_w^2 \|y_w^k - x_w^*\|^2. \nonumber
			\end{align}
		
			\vspace{-17pt}
			\noindent where the second inequality is because of the fact $\E\|a - \E a\|^2 = \E\|a\|^2 - \| \E a\|^2 \leq \E\|a\|^2$ and the last inequality comes from Assumption \ref{assump:lipschitz continuous gradients}.
			
			Substituting \eqref{proof:eq4} into \eqref{proof:eq3},  we have
			\vspace{-17pt}
			\begin{align}\label{proof:eq5}
				&\E_k\|g_w^k + \lambda \sum_{v \in \mathcal{R}_w} sign(x_w^k - x_v^k) + \lambda \sum_{v \in \mathcal{B}_w} sign(x_w^k - z_v^k)\|^2 \nonumber \\
				\leq& 8\E_k\|F'_{w}(x_w^k) - F'_{w}(x_w^*)\|^2  + 32\lambda^2|\mathcal{R}_w|^2p + 4\lambda^2|\mathcal{B}_w|^2p + 4L_w^2 \|x_w^k - x_w^*\|^2 + 4L_w^2 \|y_w^k - x_w^*\|^2.
			\end{align}

				\vspace{-17pt}
			For the third term at the right-hand side of \eqref{proof:eq1}, substituting the optimality condition \eqref{proof:eq2}, we have
				\vspace{-17pt}
			\begin{align}\label{proof:eq6}
				&-2\langle \E_k[g_w^k] + \lambda \sum_{v \in \mathcal{R}_w} sign(x_w^k - x_v^k), x_w^k - x_w^* \rangle \\
				=&-2\langle  F'_w(x_w^k) + \lambda \sum_{v \in \mathcal{R}_w} sign(x_w^k - x_v^k)  \quad - F'_w(x_w^*) - \lambda \sum_{v \in \mathcal{R}_w} s_{wv}, x_w^k - x_w^* \rangle\nonumber\\
				=&-2\langle F'_w(x_w^k) - F'_w(x_w^*), x_w^k - x_w^* \rangle- 2 \langle \lambda \sum_{v \in \mathcal{R}_w} sign(x_w^k - x_v^k) - \lambda \sum_{v \in \mathcal{R}_w} s_{wv}, x_w^k - x_w^* \rangle \nonumber\\
				\leq& -\frac{2\mu_w L_w}{\mu_w + L_w} \|x_w^k - x_w^*\|^2 - \frac{2}{\mu_w + L_w} \|F'_w(x_w^k) - F'_w(x_w^*)\| ^2  - 2 \langle \lambda \sum_{v \in \mathcal{R}_w} sign(x_w^k - x_v^k) - \lambda \sum_{v \in \mathcal{R}_w} s_{wv}, x_w^k - x_w^* \rangle. \nonumber
			\end{align}	

			\vspace{-17pt}
			\noindent where the last inequality comes from \cite{nesterov2003introductory} since we assume that the functions $F_w(x_w)$ are strongly convex and have Lipschitz continuous gradients (cf. Assumptions \ref{assump:strong convexity} and \ref{assump:lipschitz continuous gradients}).
			
			For the last term at the right-hand side of \eqref{proof:eq1}, with any $\epsilon>0$ we have
				\vspace{-17pt}
			\begin{align}\label{proof:eq7}
				&-2\langle\lambda \sum_{v \in \mathcal{B}_w} sign(x_w^k - z_v^k), x_w^k - x_w^* \rangle \\
				\leq&\epsilon\|x_w^k - x_w^*\|^2 + \frac{\lambda^2}{\epsilon}\|\sum_{v \in \mathcal{B}_w} sign(x_w^k - z_v^k)\|^2 \nonumber\\
				\leq&\epsilon\|x_w^k - x_w^*\|^2 + \frac{\lambda^2|\mathcal{B}_w|^2p}{\epsilon}. \nonumber
			\end{align}

				\vspace{-17pt}
			Substituting \eqref{proof:eq5}, \eqref{proof:eq6} and \eqref{proof:eq7} into \eqref{proof:eq1} and combining the terms, we have
			\vspace{-17pt}
			\begin{align}\label{proof:eq8}
				&\E_k \|x_w^{k+1} - x_w^*\|^2 \\
				\leq& (1 - \alpha(\frac{2\mu_w L_w}{\mu_w + L_w } - \epsilon) + 4\alpha^2L_w^2) \|x_w^k - x_w^*\|^2 + 4\alpha^2L_w^2 \|y_w^k - x_w^*\|^2 + \alpha^2 (32\lambda^2|\mathcal{R}_w|^2p + 4\lambda^2|\mathcal{B}_w|^2p) + \alpha \frac{\lambda^2|\mathcal{B}_w|^2p}{\epsilon} \nonumber\\
				&- 2 \langle \lambda \sum_{v \in \mathcal{R}_w} sign(x_w^k - x_v^k) - \lambda \sum_{v \in \mathcal{R}_w} s_{wv}, x_w^k - x_w^*\rangle - 2\alpha (\frac{1}{\mu_w + L_w} - 4\alpha) \|F'_w(x_w^k) - F'_w(x_w^*)\|^2.  \nonumber
			\end{align}

				\vspace{-17pt}
			If we constrain the step size as
			\vspace{-17pt}
			\begin{align}\label{proof:step_size_rule_1}
				\alpha \leq \min_{w \in \mathcal{R}} \{\frac{1}{4(\mu_w + L_w)}\},
			\end{align}

			\vspace{-17pt}
			\noindent we have $\frac{1}{\mu_w + L_w} - 4\alpha \geq 0$ and hence can drop the last term at the right-hand side of \eqref{proof:eq8}. Also noticing the definitions of $\eta$ and $L$, we can rewrite \eqref{proof:eq8} as
			\vspace{-17pt}
			\begin{align}\label{proof:eq9}
				&\E_k \|x_w^{k+1} - x_w^*\|^2 \\
				\leq& (1 - 2\eta\alpha + 4\alpha^2L^2) \|x_w^k - x_w^*\|^2 + 4\alpha^2L^2 \E_k \|y_w^k - x_w^*\|^2  + \alpha^2 (32\lambda^2|\mathcal{R}_w|^2p + 4\lambda^2|\mathcal{B}_w|^2p) \nonumber\\
				&+ \alpha \frac{\lambda^2|\mathcal{B}_w|^2p}{\epsilon}  - 2 \langle \lambda \sum_{v \in \mathcal{R}_w} sign(x_w^k - x_v^k) - \lambda \sum_{v \in \mathcal{R}_w} s_{wv}, x_w^k - x_w^*\rangle. \nonumber
			\end{align}

				\vspace{-17pt}
			\textbf{Step 2.} Here we define $I_p(x) = \frac{\lambda}{2} \sum_{w \in \mathcal{R}} \sum_{v \in \mathcal{R}_w} \|x_w - x_v\|_1$. Since $I_p(x)$ is convex, we have
				\vspace{-17pt}
			\begin{align}\label{proof:eq10}
				\langle \partial_x I_p(x^k) - \partial_x I_p(x^*), x^k - x^* \rangle
				=\sum_{w \in \mathcal{R}} \langle \lambda \sum_{v \in \mathcal{R}_w} sign(x_w^k - x_v^k) - \lambda \sum_{v \in \mathcal{R}_w} s_{wv}, x_w^k - x_w^*\rangle \geq 0.
			\end{align}

				\vspace{-17pt}
			Summing up \eqref{proof:eq9} over all regular agents $w \in \mathcal{R}$ and adding to \eqref{proof:eq10}, we have
				\vspace{-17pt}
			\begin{align}\label{proof:eq11}
				&\E_k \|x^{k+1} - x^*\|^2 \\
				\leq& (1 - 2\eta\alpha + 4\alpha^2L^2) \|x^k - x^*\|^2 + 4\alpha^2L^2\sum_{w \in \mathcal{R}} \|y_w^k - x_w^*\|^2 + \alpha^2 \sum_{w \in \mathcal{R}} (32\lambda^2|\mathcal{R}_w|^2p + 4\lambda^2|\mathcal{B}_w|^2p) + \alpha \sum_{w \in \mathcal{R}} \frac{\lambda^2|\mathcal{B}_w|^2p}{\epsilon}. \nonumber
			\end{align}

				\vspace{-17pt}
			\textbf{Step 3.} Defining $S^k := \sum_{w \in \mathcal{R}} \| y_w^k - x_w^*\|^2$, we have
				\vspace{-17pt}
			\begin{align}\label{proof:eq12}
				 \E_k [S^{k+1}]
				= \sum_{w \in \mathcal{R}}  \E_k \|y_w^{k+1} - x_w^*\|^2
				= (1 - \frac{1}{J}) \sum_{w \in \mathcal{R}} \|y_w^k - x_w^*\|^2 + \frac{1}{J} \sum_{w \in \mathcal{R}} \|x_w^k - x_w^*\|^2
				= (1 - \frac{1}{J}) S^k + \frac{1}{J} \|x^k - x^*\|^2.
			\end{align}

			Combining \eqref{proof:eq11} and \eqref{proof:eq12}, we have
				\vspace{-17pt}
			\begin{align}\label{proof:eq13}
				& \E_k \|x^{k+1} - x^*\|^2 + 8 J\alpha^2L^2 \E_k [S^{k+1}]\\
				\leq& (1 - 2\eta\alpha + 12\alpha^2L^2) \|x^k - x^*\|^2 + (1 - \frac{1}{2J}) 8J\alpha^2L^2S^k + \alpha^2 \sum_{w \in \mathcal{R}} (32\lambda^2|\mathcal{R}_w|^2p + 4\lambda^2|\mathcal{B}_w|^2p) + \alpha \sum_{w \in \mathcal{R}} \frac{\lambda^2|\mathcal{B}_w|^2p}{\epsilon}. \nonumber
			\end{align}

				\vspace{-17pt}
			If we constrain the step size as
				\vspace{-17pt}
			\begin{align}\label{proof:step_size_rule_2}
				12 \alpha^2 L^2 \leq \eta\alpha \quad \text{and} \quad  \eta\alpha \leq \frac{1}{2J},
			\end{align}

			\vspace{-17pt}
			\noindent then the coefficients in \eqref{proof:eq13} satisfy
				\vspace{-17pt}
			\begin{align}
				1- 2\eta \alpha + 12 \alpha^2 L^2 \leq 1 - \eta\alpha, \\
				1 - \frac{1}{2J} \leq 1 - \eta\alpha.
			\end{align}

				\vspace{-17pt}
			Construct a Lyapunov function as $V^k := \|x^k - x^*\|^2 + 8J\alpha^2L^2S^k$, we have
				\vspace{-17pt}
			\begin{align}\label{proof:eq14}
				 \E_k[V^k]
				\leq (1 - \eta\alpha) V^{k - 1} + \alpha^2 \sum_{w \in \mathcal{R}} (32\lambda^2|\mathcal{R}_w|^2p + 4\lambda^2|\mathcal{B}_w|^2p) + \alpha \sum_{w \in \mathcal{R}} \frac{\lambda^2|\mathcal{B}_w|^2p}{\epsilon}.
			\end{align}

			\vspace{-17pt}
			Taking the full expectation of \eqref{proof:eq14}, we obtain
			\vspace{-17pt}
			\begin{align}\label{proof:eq14-2}
				\E[V^k]
				\leq (1 - \eta\alpha) \E [V^{k - 1}] + \alpha^2 \sum_{w \in \mathcal{R}} (32\lambda^2|\mathcal{R}_w|^2p + 4\lambda^2|\mathcal{B}_w|^2p) + \alpha \sum_{w \in \mathcal{R}} \frac{\lambda^2|\mathcal{B}_w|^2p}{\epsilon}.
			\end{align}

			\vspace{-17pt}
			Using telescopic cancellation on \eqref{proof:eq14-2} from time 1 to time $k$, we have
			\vspace{-17pt}
			\begin{align}\label{proof:eq15}
				\E[V^k] \leq (1 - \eta\alpha)^k V^0 + \Delta_1,
			\end{align}

			\vspace{-17pt}
			\noindent where
			\vspace{-17pt}
			\begin{align}
				\Delta :=  \frac{\alpha}{\eta}\sum_{w \in \mathcal{R}} (32\lambda^2|\mathcal{R}_w|^2p + 4\lambda^2|\mathcal{B}_w|^2p)
				 + \frac{1}{\epsilon\eta}\sum_{w \in \mathcal{R}}\lambda^2|\mathcal{B}_w|^2 p.
			\end{align}

			\vspace{-17pt}
			In our derivation so far, the constraint on the step size $\alpha$ (cf. \eqref{proof:step_size_rule_1} and \eqref{proof:step_size_rule_2}) is
			\vspace{-17pt}
			\begin{align}
				\alpha \leq \min\{\min_{w \in \mathcal{R}} \{\frac{1}{4(\mu_w + L_w)}\}, \frac{\eta}{12L^2}, \frac{1}{2\eta J}\}.
			\end{align}

			Therefore, we simply choose
			\vspace{-17pt}
			\begin{align}
				\alpha \leq \frac{\eta}{12 L^2 J},
			\end{align}
			which completes the proof for BRAVO-LSVRG.
			
			\textbf{Step 4.} Below we prove Theorem 1 for BRAVO-SAGA. Since the update rules of BRAVO-SAGA and BRAVO-LSVRG are both \eqref{eq-DECEMBER-VR} and the difference between them is the definition of $g_w^k$, the intermediate results \eqref{proof:eq1}, \eqref{proof:eq3}, \eqref{proof:eq6}, and \eqref{proof:eq7} also hold true for BRAVO-SAGA. Therefore, we just need to prove \eqref{proof:eq4} and \eqref{proof:eq5} for BRAVO-SAGA.
			
			From the definition of $g_w^k$ in \eqref{eq-saga}, at every regular agent $w$, we have
			\vspace{-17pt}
			\begin{align}\label{proof-saga:eq1}
				&\E_k \|g_w^k - F'_w(x_w^k)\|^2 \\
				=&\E_k\|F'_{w, i_w^k}(x_w^k) - F'_{w, i_w^k} (x_w^*) + F'_{w}(x_w^*) - F'_w(x_w^k)
				- \Big(F'_{w, i_w^k}(\phi_{w, i_w^k}^k) - F'_{w, i_w^k}(x_w^*) + F'_w(x_w^*) - \frac{1}{J} \sum_{j=1}^J F'_{w, j}(\phi_{w, j}^k) \Big) \|^2 \nonumber\\
				\leq&2\E_k\|F'_{w, i_w^k}(x_w^k) - F'_{w, i_w^k} (x_w^*) + F'_{w}(x_w^*) - F'_w(x_w^k)\|^2  + 2\E_k\|F'_{w, i_w^k}(\phi_{w, i_w^k}^k) - F'_{w, i_w^k}(x_w^*) +  F'_{w}(x_w^*) - \frac{1}{J} \sum_{j=1}^J F'_{w, j}(\phi_{w, j}^k)\|^2 \nonumber\\
				\leq&2\E_k \|F'_{w, i_w^k}(x_w^k) - F'_{w, i_w^k} (x_w^*)\|^2  + 2\E_k\|F'_{w, i_w^k}(x_w^*) - F'_{w, i_w^k}(\phi_{w, i_w^k}^k)\|^2 \nonumber\\
				\leq&2L_w^2  \|x_w^k - x_w^*\|^2 + 2L_w^2 \E_k  \|\phi_{w, i_w^k}^k - x_w^* \|^2 \nonumber \\
				= &2L_w^2  \|x_w^k - x_w^*\|^2 + 2L_w^2 \frac{1}{J} \sum_{j=1}^J  \|\phi_{w, j}^k - x_w^*\|^2. \nonumber
			\end{align}

			\vspace{-17pt}
			Substituting \eqref{proof-saga:eq1} into \eqref{proof:eq3}, we have
			\vspace{-17pt}
			\begin{align}\label{proof-saga:eq2}
				&\E_k\|g_w^k + \lambda \sum_{v \in \mathcal{R}_w} sign(x_w^k - x_v^k) + \lambda \sum_{v \in \mathcal{B}_w} sign(x_w^k - z_v^k)\|^2 \nonumber \\
				\leq& 8\E_k\|F'_{w}(x_w^k) - F'_{w}(x_w^*)\|^2 + 4L_w^2 \frac{1}{J} \sum_{j=1}^J  \|\phi_{w, j}^k - x_w^*\|^2  + 4L_w^2  \|x_w^k - x_w^*\|^2 + 32\lambda^2|\mathcal{R}_w|^2p + 4\lambda^2|\mathcal{B}_w|^2p.
			\end{align}

			\vspace{-17pt}
			Similar to the proof for BRAVO-LSVRG, we can get
			\vspace{-17pt}
			\begin{align}\label{proof-saga:eq3}
				&\E_k \|x^{k+1} - x^*\|^2 \\
				\leq& (1 - 2\eta\alpha + 4\alpha^2L^2) \|x^k - x^*\|^2 + 4\alpha^2L^2\sum_{w \in \mathcal{R}} \frac{1}{J} \sum_{j=1}^J  \|\phi_{w, j}^k - x_w^*\|^2  + \alpha^2 \sum_{w \in \mathcal{R}} (32\lambda^2|\mathcal{R}_w|^2p + 4\lambda^2|\mathcal{B}_w|^2p) + \alpha \sum_{w \in \mathcal{R}} \frac{\lambda^2|\mathcal{B}_w|^2p}{\epsilon}. \nonumber
			\end{align}

			\vspace{-17pt}
			Define $S^k := \sum_{w \in \mathcal{R}} \frac{1}{J} \sum_{j=1}^J  \|\phi_{w, j}^k - x_w^*\|^2$, we have
			\vspace{-17pt}
			\begin{align}\label{proof-saga:eq4}
				\E_k [S^{k+1}]
				= \sum_{w \in \mathcal{R}} \frac{1}{J} \sum_{j=1}^J \E_k \|\phi_{w, j}^{k+1} - x_w^*\|^2
				=(1 - \frac{1}{J}) \sum_{w \in \mathcal{R}} \frac{1}{J} \sum_{j=1}^J \|\phi_{w, j}^k - x_w^*\|^2 + \frac{1}{J} \sum_{w \in \mathcal{R}} \|x_w^k - x_w^*\|^2
				=(1 - \frac{1}{J}) S^k + \frac{1}{J}  \|x^k - x^*\|^2.
			\end{align}
			\vspace{-17pt}
			
			With \eqref{proof-saga:eq4}, the rest of the proof follows Step 3 and the proof for BRAVO-SAGA is complete.
		\end{proof}

\section{Proof of Theorem 2}
				
\begin{proof}
We establish the lower bound on the learning error by constructing a case that the regular agents' models converge to an unsatisfactory point due to the misleading of the Byzantine agents. We construct a network topology within which the regular network composed of the regular agents is connected and each regular agent has the same number of Byzantine neighbors, namely, $|\mathcal{B}_w| = |\mathcal{B}_v|$ for any $w, v \in \mathcal{R}$.

We begin with the scalar case with $p = 1$. For each regular agent $w \in \mathcal{R}$, we construct its local cost function as
\vspace{-17pt}
\begin{align}
	F_w(\tilde{x}) = \frac{1}{2} \tilde{x}^2 - \lambda |\mathcal{B}_w| \tilde{x},
\end{align}

\vspace{-17pt}
\noindent which is strongly convex. Define the local sample costs as
\vspace{-17pt}
\begin{align}\label{proof-lower-bound: sample-cost}
	F_{w, j}(\tilde{x}) = F_w(\tilde{x}), \quad  \forall j \in \{1, \cdots, J\},
\end{align}

\vspace{-17pt}
\noindent which have Lipschitz continuous gradients with constants 1. It is easy to verify that the optimal solution of \eqref{problem-TV} is $x^* = [\lambda |\mathcal{B}_1|;\lambda |\mathcal{B}_2|; \cdots; \lambda |\mathcal{B}_R|]$ in this case.
					
Since the regular agents' models are initialized at the same point, without loss of generality, we let $x_w^0 = 0, \forall w \in \mathcal{R}$. Otherwise, we can set $F_w(\tilde{x} - x_w^0)$ as the local cost of agent $w$. Note that for each regular agent $w \in \mathcal{R} $, $h_w^0$ is a linear combination of $\{F'_{w, 1}(x_w^0), \cdots, F'_{w, J}(x_w^0)\}$. With \eqref{proof-lower-bound: sample-cost}, we have $h_w^0 \in \text{span} \{F'_w(x_w^0)\}$. Because $\E[h_w^0] =  F'_w(x_w^0)$, we have
\vspace{-17pt}
\begin{align}
	h_w^0 = F'_w(x_w^0) = -\lambda |\mathcal{B}_w|.
\end{align}

\vspace{-17pt}				
Therefore, for each regular agent $w \in \mathcal{R}$, we have
\vspace{-17pt}
\begin{align}\label{eq-proof-lowerbound-1}
	\hat{h}_w^0 &= h_w^0  + \lambda \sum_{v \in \mathcal{R}_w} sign(x_w^0 - x_v^0)
	+ \lambda \sum_{v \in \mathcal{B}_w}sign(x_w^0 - z_v^0)  \\ & = -\lambda |\mathcal{B}_w| + \lambda |\mathcal{R}_w| sign(0) + \lambda \sum_{v \in \mathcal{B}_w}sign( - z_v^0).  \nonumber
\end{align}

\vspace{-17pt}
Although $sign(0)$ could be any value in $ [-1, 1]$, we let $sign(0) = 0$ by convention. Choosing the Byzantine attacks such that $ \sum_{v \in \mathcal{B}_w}sign( - z_v^0) = |\mathcal{B}_w|$ leads to
\vspace{-17pt}
	\begin{align}
		\hat{h}_w^0 = 0.
	\end{align}		

\vspace{-17pt}				
With Assumption \ref{assump:span condition}, we further have
\vspace{-17pt}
\begin{align}
	x_w^1 \in x_w^0 + \text{span}\{\hat{h}_w^0\} = \{0\}.
\end{align}

\vspace{-17pt}
At time $k=1$, $h_w^1$ is linear combination of $\{F'_{w,1}(x_w^0), \cdots, F'_{w, J}(x_w^0), F'_{w, 1}(x_w^1), \cdots, F'_{w, J}(x_w^1)\}$ for each regular agent $w \in \mathcal{R}$. With \eqref{proof-lower-bound: sample-cost}, we have $h_w^1 \in \text{span} \{F'_w(x_w^0), F'_w(x_w^1)\} = \text{span} \{F'_w(0)\}$. Since $\E[h_w^1] = F'_w(x_w^1) = F'_w(0)$, we have
\vspace{-17pt}
\begin{align}
	h_w^1 = F'_w(0) = -\lambda |\mathcal{B}_w|.
\end{align}

\vspace{-17pt}		
Therefore, for each regular agent $w \in \mathcal{R}$, we have
\vspace{-17pt}
\begin{align}\label{eq-proof-lowerbound-2}
	\hat{h}_w^1 &= h_w^1  + \lambda \sum_{v \in \mathcal{R}_w} sign(x_w^1- x_v^1)+ \lambda \sum_{v \in \mathcal{B}_w}sign(x_w^1 - z_v^1) \nonumber \\ & = -\lambda |\mathcal{B}_w| + \lambda \sum_{v \in \mathcal{R}_w} sign(x_w^1- x_v^1) + \lambda \sum_{v \in \mathcal{B}_w} sign(- z_v^1).  \nonumber
\end{align}

\vspace{-17pt}
Choosing  the Byzantine attacks such that $\sum_{v \in \mathcal{B}_w}sign(- z_v^1) = |\mathcal{B}_w|$, we have
\vspace{-17pt}
\begin{align}
	\hat{h}_w^1 = 0.
\end{align}

\vspace{-17pt}
With Assumption \ref{assump:span condition}, we further have
\vspace{-17pt}
\begin{align}
x_w^2 \in x_w^0 + \text{span}\{\hat{h}_w^0, \hat{h}_w^1\} = \{0\}.
\end{align}

\vspace{-17pt}					
As such, for each regular agent $w \in \mathcal{R}$, we have
\vspace{-17pt}
\begin{align}
	x_w^k = 0 \quad \text{and} \quad \E \|x_w^{k} - x_w^*\|  =\|x_w^k - x_w^*\| =\lambda |\mathcal{B}_w|, \quad \forall k \in \{0, 1, \cdots\}.
\end{align}

\vspace{-17pt}
Therefore, we have
\vspace{-17pt}
\begin{align}
	\E \| x^k - x^*\|^2 = \sum_{w \in \mathcal{R}} \E\|x_w^k - x_w^*\|^2 &= \Omega( \sum_{w \in \mathcal{R}} \lambda^2 |\mathcal{B}_w|^2).
\end{align}

\vspace{-17pt}
For the vector case with $p > 1$, we apply the constructions on the scalar case for all dimensions. Consider
\vspace{-17pt}
\begin{align}
	F_w(\tilde{x}) = \frac{1}{2} \tilde{x}^T \tilde{x} - \lambda |\mathcal{B}_w| \tilde{x}^T \bm{1},
\end{align}

\vspace{-17pt}
\noindent where $x \in \mathbb{R}^p$ and $\bm{1} \in \mathbb{R}^p$ is the all-one vector. Define
\vspace{-17pt}
\begin{align}
	F_{w, j}(\tilde{x}) = F_w(\tilde{x}), \quad  \forall j \in \{1, \cdots, J\}.
\end{align}

\vspace{-17pt}
Also let $x_w^0 = \bm{0}$ for all regular agents $w \in \mathcal{R}$ and choose the Byzantine attacks such that $\sum_{v \in \mathcal{B}_w} sign(-z_v^k) = |\mathcal{B}_w| \cdot \bm{1}$ at any time $k$. Similar to the derivations of the scalar case, we have
\vspace{-17pt}
\begin{align}
	\E \| x^k - x^*\|^2 = \Omega( \sum_{w \in \mathcal{R}}\lambda^2 |\mathcal{B}_w|^2 p),
\end{align}

\vspace{-17pt}
\noindent which completes the proof.
\end{proof}

				\bibliographystyle{elsarticle-num}
				\bibliography{refs}

\end{document}